\documentclass[pdflatex,iicol,sn-basic]{sn-jnl}

\usepackage{graphicx}%
\usepackage{multirow}%
\usepackage{amsmath,amssymb,amsfonts}%
\usepackage{amsthm}%
\usepackage{mathrsfs}%
\usepackage[title]{appendix}%
\usepackage{xcolor}%
\usepackage{textcomp}%
\usepackage{manyfoot}%
\usepackage{booktabs}%
\usepackage{pifont}
\usepackage{threeparttable}
\usepackage{bm}
\usepackage{bbm}
\usepackage{colortbl}
\usepackage[normalem]{ulem}
\usepackage{makecell}
\definecolor{lightgreen}{RGB}{34,139,34}
\definecolor{lightred}{RGB}{255,100,100}
\definecolor{cell_bisque}{rgb}{1.0, 0.89, 0.77}
\definecolor{cell_blond}{rgb}{0.98, 0.94, 0.75}
\definecolor{cell_blue}{RGB}{155, 187, 228}
\definecolor{princetonorange}{rgb}{1.0, 0.56, 0.0}
\definecolor{pinkpearl}{rgb}{0.91, 0.67, 0.81}
\definecolor{mossgreen}{rgb}{0.68, 0.87, 0.68}
\definecolor{citecolor}{RGB}{0,102,204}
\definecolor{refcolor}{RGB}{218, 47, 138}
\definecolor{skyblue}{RGB}{181, 212, 244}

\theoremstyle{thmstyleone}%
\theoremstyle{thmstyletwo}%

\theoremstyle{thmstylethree}%

\begin{document}

\title[DNA: Dual-stage Native Attribution]
{DNA: Dual-stage Native Attribution for Generated Image Source Tracing}


\author[1]{\fnm{Chao} \sur{Wang}}\email{chaowang0708@mail.ustc.edu.cn}
\author*[1]{\fnm{Kejiang} \sur{Chen}}\email{chenkj@ustc.edu.cn}
\author[1]{\fnm{Zijin} \sur{Yang}}\email{bsmhmmlf@mail.ustc.edu.cn}
\author[2]{\fnm{Yaofei} \sur{Wang}}\email{wyf@hfut.edu.cn}
\author[3]{\fnm{Yuang} \sur{Qi}}\email{qiyuang@bza.edu.cn}
\author*[1]{\fnm{Weiming} \sur{Zhang}}\email{zhangwm@ustc.edu.cn}
\author[1]{\fnm{Nenghai} \sur{Yu}}\email{ynh@ustc.edu.cn}

\affil[1]{\orgdiv{School of Cyber Science and Technology}, \orgname{University of Science and Technology of China}, \orgaddress{\street{No. 96 Jinzhai Road, Baohe District}, \city{Hefei}, \postcode{230026}, \state{Anhui}, \country{China}}}

\affil[2]{\orgdiv{School of Computer Science and Information Engineering}, \orgname{Hefei University of Technology}, \orgaddress{\street{No. 193 Tunxi Road}, \city{Hefei},
\postcode{230009}, \state{Anhui}, \country{China}}}

\affil[3]{\orgname{Zhongguancun Academy}, \orgaddress{\city{Beijing}, \postcode{100094}, \state{Beijing}, \country{China}}}


\abstract{
The rapid evolution of image generation has produced numerous within-family variants, making source-model attribution of suspect images increasingly important for digital forensics. Proactive methods require watermarking or model modification, while passive methods often depend on extensive supervised training or a single reconstruction signal, limiting their ability to handle unknown sources and distinguish highly similar variants. We observe that attribution signals in latent generative models are naturally stratified across architectural levels: Variational autoencoder (VAE) cues reflect family-shared information, whereas backbone cues capture variant-specific behaviors. Motivated by this insight, we propose \emph{Dual-stage Native Attribution (DNA)}, a coarse-to-fine (\emph{VAE}$\rightarrow$\emph{backbone}) framework requiring no neural-network training, with supervision limited to lightweight calibration. The coarse-grained stage uses \emph{Autoencoder Double-Reconstruction (AEDR)} for efficient open-set family-level screening. The fine-grained stage performs closed-set model-level attribution with \emph{Native Prediction Consistency (NPC)}, which compares native prediction errors of within-family variants across multiple noise levels under semantic conditioning and attributes the source via normalized calibrated scores. We construct DNA-30K—to our knowledge, the first benchmark for within-family variant attribution under open-set family-level evaluation. It comprises 30,000 images: 24 candidate models across six families spanning denoising diffusion and flow matching, plus non-candidate generated and natural images as unknown sources. Experiments show that DNA achieves 89.11\% end-to-end accuracy, outperforming the strongest baseline by 33.15\% even with NPC as the fine-grained stage. DNA remains stable under limited calibration samples and diverse generation configurations.
}




\maketitle

\section{Introduction}
\label{Intro}

\begin{figure*}[t]
\centering
\includegraphics[width=1\textwidth]{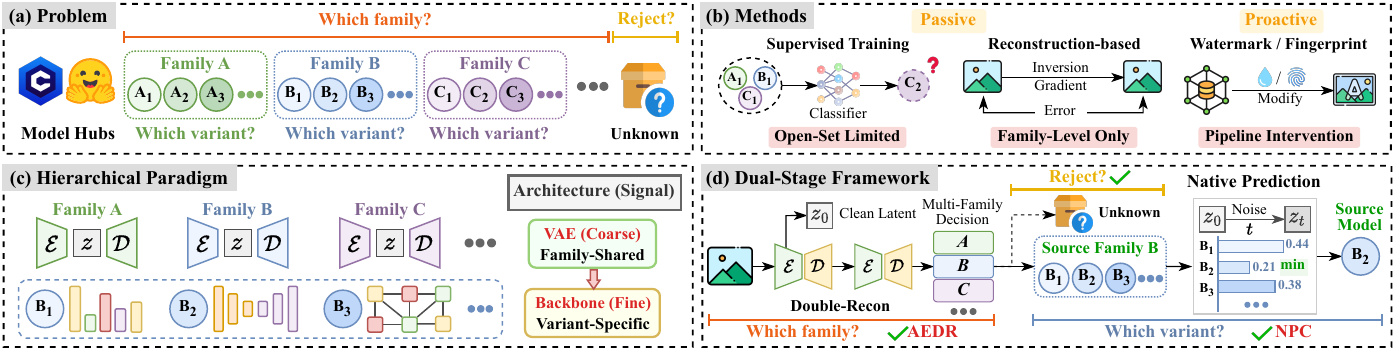}
\caption{Problem setting and the proposed paradigm. (a) Within-family variant attribution under open-set family-level conditions; (b) Limitations of prior approaches; (c) The coarse-to-fine hierarchical paradigm: VAE-level cues carry family-shared information, backbone-level cues carry variant-specific information; (d) The proposed dual-stage DNA framework.}
\label{Problem_Methods}
\end{figure*}

Denoising diffusion~\citep{DDPM} and flow matching~\citep{FM} have become the dominant paradigms in image generation, producing high-quality content from simple prompts~\citep{LDM}. As photorealism and semantic alignment continue to improve, generated imagery increasingly blurs the boundary between synthetic and authentic content, permeating creative design, film production, and other visual applications~\citep{DM-use}.

However, the democratization of image generation has also amplified misuse risks, as exemplified by fabricated Pentagon-explosion imagery~\citep{Pentagon}. Merely detecting whether an image is AI-generated is insufficient for accountability; reliable provenance requires tracing a suspect image back to the source model that produced it. This need is sharpened by the rapid expansion of open-weight generative ecosystems\footnote{\url{https://huggingface.co}}\footnote{\url{https://civitai.com}}: public model hubs now host many checkpoints that are fine-tuned, distilled, or otherwise derived from a small set of shared base models (Figure~\ref{Problem_Methods}). In such ecosystems, intellectual-property protection, license compliance, and responsibility attribution depend on identifying not merely the model family, but the exact within-family variant. Because these candidate models are themselves openly accessible, fine-grained attribution among highly similar within-family variants becomes a practically tractable yet still unsolved problem.

Existing attribution methods for generated images can be broadly divided into proactive and passive approaches (Figure~\ref{Problem_Methods}). Proactive methods embed traceable signals, such as watermarks~\citep{HiDDeN, FIN, MBRS, PIMoG, GS, GS++, Tree-Ring, Signature} or model fingerprints~\citep{Finger-1, Finger-2, Finger2DM, Finger2Sem}, during training or generation to establish ownership identifiers. However, they typically require intervening in model weights or generation pipelines, which may constrain deployment or compromise visual quality. Passive methods instead perform post-hoc attribution from image-intrinsic evidence, without prior intervention. Supervised passive methods~\citep{GANFingerprints-Data, CNN-Res-ForenSynths-Data, DE-FAKE, UniversalAttribution, CDAL, ZBH} train on large-scale generated images, but often generalize poorly to non-candidate data and require retraining for new sources. More critically, the highly similar generative distributions of within-family variants make discriminative boundaries difficult to learn~\citep{WACV}. Reconstruction-based methods~\citep{RONAN, LatentTracer, AEDR} instead exploit an intrinsic property: a source model exhibits stronger reconstruction or prediction consistency for images from its own distribution. This provides an attribution signal compatible with open-set family-level scenarios, making such methods promising for real-world forensics.

\begin{figure*}[t]
\centering
\includegraphics[width=1\textwidth]{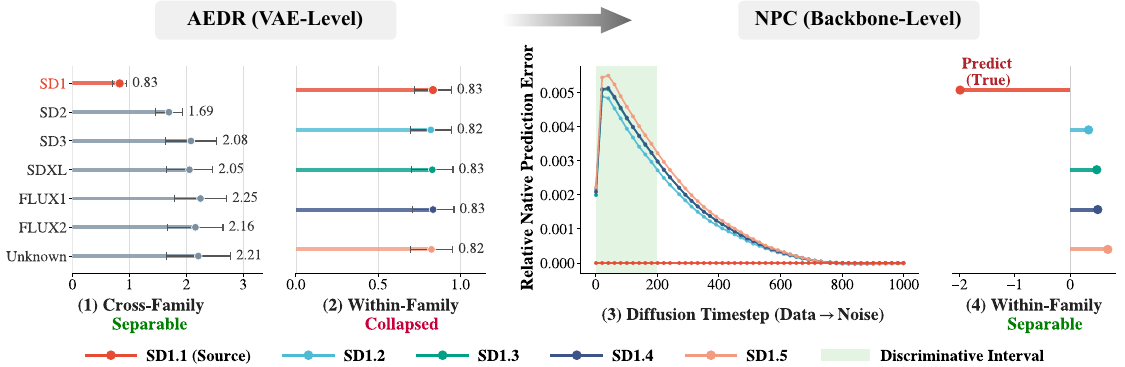}
\caption{Attribution signals at two architectural levels, using the Stable Diffusion v1 family as an example. \textit{Column (1):} the VAE-level AEDR signal is highly separable in the cross-family setting, enabling coarse-grained family-level attribution. \textit{Column (2):} the signal collapses among within-family variants. \textit{Column (3):} the backbone-level NPC signal shows the relative native prediction error across noise/time levels, where larger separation between the source and non-source curves indicates stronger discriminability, and the shaded region denotes the discriminative interval. \textit{Column (4):} after z-score normalization and aggregation over this interval, the true source attains the minimum score.}
\label{AEDR-NPC-All}
\end{figure*}

Within reconstruction-based attribution for latent generative models, LatentTracer~\citep{LatentTracer} and AEDR~\citep{AEDR} are two notable approaches. LatentTracer measures the reconstruction capability of candidate models through latent-space inversion and gradient-based optimization~\citep{Adam}, using reconstruction error as the attribution signal. Our prior work, AEDR, replaces a single reconstruction error with a homogeneity-calibrated ratio of two cascaded variational autoencoder (VAE) reconstructions, yielding a more stable signal at substantially lower computational cost. However, AEDR derives its signal from VAE-level reconstruction behavior: when models within a family share identical or highly compatible VAEs, this signal cannot capture the subtle differences among within-family variants (Figure~\ref{AEDR-NPC-All}). This raises a central question: \emph{if VAE-level signals only reveal family-level membership, which component provides fine-grained model-level evidence?}

This question prompts us to revisit the internal structure of latent generative models. We observe an underexplored structural cue (Figure~\ref{Problem_Methods}): \textbf{\emph{attribution signals are naturally organized along the architectural hierarchy}}. Specifically, the VAE maps images between pixel and latent spaces~\citep{VAE}, defining the latent distribution on which the generative backbone operates; the backbone in turn learns the denoising diffusion or flow matching process within this latent space. Because the two components are tightly coupled, within-family variants often share identical or highly compatible VAEs while their variant-specific differences mainly reside in the backbone. This hierarchy naturally separates attribution signals: VAE-level cues reveal family-level membership for coarse attribution, whereas backbone-level cues capture variant-specific behaviors for fine attribution.

Building on this insight, we decouple the attribution task along the ``\emph{VAE$\rightarrow$backbone}'' hierarchy into a ``\emph{coarse$\rightarrow$fine}'' pipeline, aligning the attribution granularity with the signals carried by each architectural level. We propose \textit{\textbf{Dual-stage Native Attribution (DNA)}}, a coarse-to-fine attribution framework comprising Autoencoder Double-Reconstruction (\textit{\textbf{AEDR})}~\citep{AEDR} at the coarse-grained stage and Native Prediction Consistency (\textit{\textbf{NPC}}) at the fine-grained stage (Figure~\ref{Problem_Methods}). The former leverages family-shared VAE-level signals to rapidly narrow the search space from all candidate models to a family, while the latter exploits variant-specific backbone-level signals to distinguish within-family variants. Without additional neural-network training, DNA enables end-to-end attribution from family-level identification to exact source-model localization, combining efficient coarse-grained screening with precise fine-grained discrimination.

The coarse-grained stage builds on our prior work AEDR, using the homogeneity-calibrated VAE double-reconstruction ratio as a family-level signal to narrow the search space to a unique candidate family or reject unknown sources. The fine-grained stage introduces NPC, which exploits the native prediction consistency of the generative backbone: a source model is expected to produce lower and more stable prediction errors on perturbed latents derived from its belonging images (Figure~\ref{AEDR-NPC-All}). NPC evaluates this consistency across multiple noise levels and maps the resulting errors into a normalized and calibrated score space. This formulation naturally accommodates both denoising diffusion and flow matching, enabling closed-set model-level attribution.

To support systematic evaluation of the proposed framework, we construct DNA-30K. It comprises 30,000 images from 24 candidate models across six model families, spanning both denoising diffusion and flow matching, together with images from non-candidate generators and natural-image datasets as unknown sources. Existing benchmarks~\citep{Attribution88-Data, IFDL-Data, OSMA-Data, GenImage-Data, WildFake-Data, WILD-Data} primarily focus on coarse-grained source identification across architectures or model families, with limited coverage of within-family variants and recently emerged flow-matching models. To the best of our knowledge, DNA-30K is the first benchmark designed to jointly support open-set family-level screening and closed-set within-family variant attribution.

Our contributions are summarized as follows:

\begin{itemize}
\item \textbf{Hierarchical Insight.} We reveal that attribution cues in latent generative models are stratified across architectural levels: the VAE mainly preserves family-level reconstruction signatures, while the backbone captures finer model-level prediction behaviors. Based on this, we propose a ``\emph{coarse$\rightarrow$fine}'' decoupled paradigm along the ``\emph{VAE}$\rightarrow$\emph{backbone}'', formulating attribution as open-set family-level screening followed by closed-set within-family variant identification.

\item \textbf{Dual-Stage Framework.} 
We propose DNA, comprising AEDR for open-set family-level screening and the newly introduced NPC for closed-set model-level attribution. NPC evaluates native prediction consistency across multiple noise levels under semantic conditioning and produces normalized and calibrated scores, supporting denoising diffusion and flow matching.

\item \textbf{Open-Set Benchmark.} 
We construct DNA-30K, to our knowledge, the first benchmark to jointly support open-set family-level screening and closed-set within-family variant attribution. It comprises 30{,}000 images from 24 candidate models across six families spanning both denoising diffusion and flow matching, and further includes unknown sources.

\item \textbf{Systematic Evaluation.}
Extensive experiments on DNA-30K show that DNA achieves 89.11\% end-to-end accuracy and outperforms the strongest baseline by 33.15\%, even when NPC is used as the fine-grained model-level attribution stage. DNA also remains stable with limited calibration samples, diverse sampling configurations, and candidate-set expansion.
\end{itemize}

\noindent
\textbf{Extension Statement.} This work substantially extends our conference paper AEDR~\citep{AEDR}. The conference version focused on VAE-level family attribution. This journal version advances from family-level to model-level attribution among within-family variants. While retaining AEDR as the coarse-grained stage and augmenting it with an open-set multi-family decision mechanism, we introduce the newly proposed hierarchical attribution paradigm, the fine-grained NPC module, the DNA-30K open-set benchmark, and comprehensive evaluations.

\section{Related Work}
\label{Related}

\subsection{Latent Generative Models}
Latent generative models have become a dominant approach for high-quality image synthesis and can be categorized by their generative dynamics into two paradigms. Denoising diffusion models originate from DDPM~\citep{DDPM} and were subsequently scaled to high-resolution synthesis by Latent Diffusion Models~\citep{LDM}, which perform the diffusion process in a low-dimensional latent space defined by a pretrained VAE~\citep{VAE}. This design underlies the Stable Diffusion series~\citep{LDM, SDXL} and also supports Transformer-based diffusion architectures~\citep{DiT}, which replace the U-Net backbone~\citep{U-Net} for improved scalability. Flow matching models, represented by Flow Matching~\citep{FM} and Rectified Flow~\citep{RF}, characterize generation through continuous probability paths and velocity-field modeling, and have become an important technical foundation for large-scale models such as Stable Diffusion 3~\citep{SD3}. Both paradigms employ a pretrained VAE that defines the latent space for the backbone; Section~\ref{Preliminaries} formalizes this unified view.

\subsection{Proactive Attribution}
Proactive attribution methods identify the source model by deliberately embedding traceable signals before, during, or after image generation, mainly including image watermarking and model fingerprinting. Image watermarking methods can be further divided into post-processing watermarks~\citep{HiDDeN, FIN, MBRS, PIMoG} and generative watermarks~\citep{GS, Tree-Ring, Signature}. The former typically embed detectable signals in the pixel or frequency domain of generated images, and thus offer broad applicability. The latter integrate watermarking into the sampling process, latent representation, or decoder, aiming to achieve traceability while preserving generation quality. Representative methods include Stable Signature~\citep{Signature}, Tree-Ring~\citep{Tree-Ring}, and Gaussian Shading~\citep{GS}, which introduce verifiable identifiers by fine-tuning the latent diffusion decoder, embedding structured watermark patterns into the initial diffusion noise, or mapping watermark information into latent representations.

Model fingerprinting methods~\citep{Finger-1, Finger-2, Finger2DM, Finger2Sem} assign source-specific signatures during data preparation, model training, or generation, enabling generated images to carry identifiable model-level traces. Artificial Fingerprints~\citep{Finger-1} is a representative work, embedding fingerprints into training data and exploiting their transferability to generative models. Later studies extend this idea to diffusion models by modulating model weights~\citep{Finger2DM} or exploiting latent semantic dimensions as fingerprints~\citep{Finger2Sem}. Proactive attribution can provide provenance evidence when the generation pipeline is controllable, but it requires prior intervention and may introduce deployment constraints, overhead, or visual quality trade-offs. In contrast, passive attribution requires no pipeline modification and performs post-hoc source tracing from image-intrinsic features or model responses, making it more applicable to real-world scenarios.

\begin{table*}[t]
\centering
\caption{Comparison of existing attribution benchmarks. ``\textcolor{princetonorange}{\ensuremath{\bm{\triangle}}}'' indicates partial coverage.}
\label{Compare_Benchmark}
\scriptsize
\setlength{\tabcolsep}{5pt}
\renewcommand{\arraystretch}{1.0}
\begin{tabular}{@{}c@{}}
\resizebox{\textwidth}{!}{%
\begin{tabular}{lccccccc}
\toprule
\textbf{Benchmark} & \textbf{Models} & \textbf{Families} & \textbf{Variants} & \textbf{Diffusion} & \textbf{Flow-Matching} & \textbf{Open-Set} & \textbf{Size} \\
\midrule
ForenSynths~\citep{CNN-Res-ForenSynths-Data}
& 6 & -- & \textcolor{lightred}{\ding{55}} & \textcolor{lightred}{\ding{55}} & \textcolor{lightred}{\ding{55}} & \textcolor{lightred}{\ding{55}} & 72.4K \\
Attribution88~\citep{Attribution88-Data}
& 7 & -- & \textcolor{lightred}{\ding{55}} & \textcolor{lightred}{\ding{55}} & \textcolor{lightred}{\ding{55}} & \textcolor{lightred}{\ding{55}} & 1.06M \\
IFDL~\citep{IFDL-Data}
& 10 & -- & \textcolor{lightred}{\ding{55}} & \textcolor{lightgreen}{\ding{51}} & \textcolor{lightred}{\ding{55}} & \textcolor{lightred}{\ding{55}} & 1.9M \\
OSMA~\citep{OSMA-Data}
& 67 & -- & \textcolor{lightred}{\ding{55}} & \textcolor{lightred}{\ding{55}} & \textcolor{lightred}{\ding{55}} & \textcolor{lightgreen}{\ding{51}} & 300K \\
De-Fake~\citep{DE-FAKE}
& 4 & -- & \textcolor{lightred}{\ding{55}} & \textcolor{lightgreen}{\ding{51}} & \textcolor{lightred}{\ding{55}} & \textcolor{lightred}{\ding{55}} & 80K \\
GenImage~\citep{GenImage-Data}
& 8 & -- & \textcolor{princetonorange}{\ensuremath{\bm{\triangle}}} & \textcolor{lightgreen}{\ding{51}} & \textcolor{lightred}{\ding{55}} & \textcolor{lightred}{\ding{55}} & 2.68M \\
WildFake~\citep{WildFake-Data}
& 21 & -- & \textcolor{princetonorange}{\ensuremath{\bm{\triangle}}} & \textcolor{lightgreen}{\ding{51}} & \textcolor{lightred}{\ding{55}} & \textcolor{lightred}{\ding{55}} & 3.57M \\
WILD~\citep{WILD-Data}
& 20 & -- & \textcolor{princetonorange}{\ensuremath{\bm{\triangle}}} & \textcolor{lightgreen}{\ding{51}} & \textcolor{lightgreen}{\ding{51}} & \textcolor{lightgreen}{\ding{51}} & 50K \\
\midrule
\textbf{DNA-30K}
& \textbf{24} & \textbf{6} & \textcolor{lightgreen}{\ding{51}} & \textcolor{lightgreen}{\ding{51}} & \textcolor{lightgreen}{\ding{51}} & \textcolor{lightgreen}{\ding{51}} & \textbf{30K} \\
\bottomrule
\end{tabular}%
}
\end{tabular}
\end{table*}

\subsection{Passive Attribution}
One line of passive attribution methods relies on supervised learning~\citep{CNN-Res-ForenSynths-Data, GANFingerprints-Data, WACV, UniversalAttribution}, training discriminative models on multi-source images to capture source-specific statistical traces. Early studies showed that CNN-generated images often contain frequency artifacts induced by upsampling~\citep{Fingerprints}, indicating that generative models leave learnable traces. Yu et al. extended this idea from real/fake detection to multi-class source identification~\citep{GANFingerprints-Data}. In the diffusion era, De-Fake~\citep{DE-FAKE} trains classifiers on CLIP~\citep{CLIP} features to identify images from multiple diffusion sources, while OCC-CLIP~\citep{OCC-CLIP} formulates attribution as a few-shot one-class classification problem, learning prompt vectors to distinguish belonging from non-belonging images. Recent works explore open-set attribution~\citep{Forensic, CDAL}. While these methods have demonstrated strong performance under supervised attribution settings, their reliance on coverage of training sources makes generalization to new sources and fine-grained within-family variants less straightforward.

Reconstruction-based methods form another line of passive attribution. Rather than relying on additional training, these methods exploit an intrinsic property of generative models: a source model typically exhibits stronger reconstruction or prediction consistency for images from its own distribution. RONAN~\citep{RONAN} is an early representative method, performing gradient-based inversion from random initialization and using complexity-calibrated reconstruction errors as attribution signals. LatentTracer~\citep{LatentTracer} transfers this idea to the latent space, conducting inversion over VAE-encoded latent variables to reduce attribution cost and improve applicability to latent generative models. Our prior work AEDR~\citep{AEDR} replaces a single absolute reconstruction error with the ratio of two cascaded VAE reconstruction errors, and further calibrates this signal with image homogeneity to suppress content-complexity bias. This design improves the stability and discriminability of reconstruction-based attribution signals. Existing reconstruction-based methods typically exploit coarse signals from a single architectural level, making them more suitable for family-level screening than for discriminating highly similar within-family variants. Consequently, within-family variant attribution remains underexplored.

\subsection{Image Attribution Benchmarks}
With the development of generated image attribution, various forensic and attribution benchmarks have been constructed. Early benchmarks mainly focused on GAN-generated image detection and attribution, such as ForenSynths~\citep{CNN-Res-ForenSynths-Data}, GAN Fingerprints~\citep{GANFingerprints-Data}, and Attribution88~\citep{Attribution88-Data}, which advanced this field from the perspectives of cross-generator generalization, model-fingerprint learning, and robust attribution representation. Later studies further extended benchmark construction to hierarchical image forensics, open-set model attribution, and synthetic-image association in real-world web scenarios, including IFDL~\citep{IFDL-Data}, OSMA~\citep{OSMA-Data}, and WILD~\citep{WILD-Data}. Recent benchmarks such as GenImage~\citep{GenImage-Data} and WildFake~\citep{WildFake-Data} cover broader types of generative models, while WILD also includes DiT-based generators, providing important evaluation resources for source analysis across architectures.

As shown in Table~\ref{Compare_Benchmark}, existing benchmarks have advanced generated-image detection and attribution across architectures, model families, and real-world scenarios. However, they provide limited support for fine-grained attribution among within-family variants. Such variants often share backbone structures, training data, or latent representation spaces, making their differences subtler than those across architectures or product-level models. This setting is therefore more challenging and closer to practical forensic demands. Meanwhile, public benchmarks for flow matching models such as Stable Diffusion 3 and FLUX, especially their within-family variants, remain limited. To fill this gap, we construct DNA-30K as a benchmark for within-family variant attribution under open-set family-level evaluation.

\section{Preliminaries}
\label{Preliminaries}

\subsection{Variational Autoencoders}
\label{VAE_Preliminaries}

In mainstream latent generative models~\citep{LDM, SDXL, SD3, FLUX1, FLUX2}, the VAE serves as a mapping component between pixel space and latent space, consisting of an encoder $\mathcal{E}$ and a decoder $\mathcal{D}$. The VAE is typically pretrained before the generative backbone and kept frozen during subsequent denoising diffusion or flow matching training, enabling generation in a low-dimensional latent space with significantly reduced computational cost. For notational simplicity, we absorb the scaling factor and its inverse into $\mathcal{E}$ and $\mathcal{D}$. 

Given an image $\mathbf{x}$, the VAE encoding and decoding are given by $\mathbf{z}_0 = \mathcal{E}(\mathbf{x})$ and $\tilde{\mathbf{x}} = \mathcal{D}(\mathbf{z}_0)$, where $\mathbf{z}_0$ denotes the clean latent and $\tilde{\mathbf{x}}$ the reconstructed image. We define the VAE reconstruction operator and reconstruction error as:
\begin{equation}
\mathcal{R}(\mathbf{x}) \triangleq \mathcal{D}(\mathcal{E}(\mathbf{x})),\quad \ell_{\mathrm{rec}}(\mathbf{x}) = d_{\mathrm{rec}}\big(\mathbf{x},\, \mathcal{R}(\mathbf{x})\big),
\label{VAE_Recon}
\end{equation}
where $d_{\mathrm{rec}}(\cdot,\cdot)$ denotes a reconstruction distance such as pixel-level $\ell_1$ or $\ell_2$. For samples aligned with the VAE training distribution, $\mathcal{R}$ approximates the identity mapping and thus yields small reconstruction errors. For samples deviating from this distribution, the VAE tends to project them toward the learned manifold, resulting in notable shifts. This asymmetric reconstruction behavior forms the basis of the family-level attribution signal in the coarse-grained stage (Section~\ref{AEDR_VAE}).

\subsection{Denoising Diffusion Models}
\label{Denoising_Diffusion_Models}

Denoising diffusion models~\citep{DDPM} generate images by learning the reverse generative dynamics of perturbed latent representations across varying noise levels. For unified notation, we adopt the common $\boldsymbol{\varepsilon}$-prediction formulation. Given an image-condition pair $(\mathbf{x},\mathbf{c})\sim p_{\mathrm{data}}$, where $p_{\mathrm{data}}$ denotes the data distribution and $\mathbf{c}$ is the semantic condition, the clean latent is obtained as $\mathbf{z}_0=\mathcal{E}(\mathbf{x})$. At a discrete timestep $t\in\{1,\dots,T\}$, where $T$ is the total number of diffusion steps, the forward process produces a perturbed latent $\mathbf{z}_t = \sqrt{\bar{\alpha}_t}\,\mathbf{z}_0 + \sqrt{1-\bar{\alpha}_t}\,\boldsymbol{\varepsilon}$, where $\boldsymbol{\varepsilon}\sim\mathcal{N}(\mathbf{0},\mathbf{I})$ denotes the injected standard Gaussian noise, and $\bar{\alpha}_t=\prod_{s=1}^{t}(1-\beta_s)$ is the cumulative noise coefficient defined by the variance schedule $\{\beta_t\}_{t=1}^{T}$. The noise predictor $\boldsymbol{\varepsilon}_{\theta}(\mathbf{z}_t,t,\mathbf{c})$, parameterized by the generative backbone parameters $\theta$, is trained to estimate $\boldsymbol{\varepsilon}$:
\begin{equation}
\mathcal{L}_{\boldsymbol{\varepsilon}}(\theta)
= \mathbb{E}\!\left[\left\|\boldsymbol{\varepsilon} - \boldsymbol{\varepsilon}_{\theta}(\mathbf{z}_t,t,\mathbf{c})\right\|_2^2\right],
\label{DD_Loss}
\end{equation}
where the expectation is taken over $(\mathbf{x},\mathbf{c})\sim p_{\mathrm{data}}$, $t\sim\mathcal{U}(\{1,\dots,T\})$, and $\boldsymbol{\varepsilon}\sim\mathcal{N}(\mathbf{0},\mathbf{I})$. This objective defines the native prediction target of denoising diffusion backbones.

\subsection{Flow Matching Models}

Flow matching models learn a velocity field along a continuous probability path~\citep{FM, SD3, FLUX1, FLUX2}. For unified notation, we present the rectified flow formulation~\citep{RF}. Given an image-condition pair $(\mathbf{x},\mathbf{c})\sim p_{\mathrm{data}}$, the clean latent is obtained as $\mathbf{z}_0=\mathcal{E}(\mathbf{x})$. At continuous time $t\in[0,1]$, the perturbed latent representation is $\mathbf{z}_t = (1-t)\,\mathbf{z}_0 + t\,\boldsymbol{\varepsilon}$, $\boldsymbol{\varepsilon}\sim\mathcal{N}(\mathbf{0},\mathbf{I})$, where $t{=}0$ corresponds to the data endpoint and $t{=}1$ to the noise endpoint. The corresponding target velocity is $d\mathbf{z}_t/dt = \boldsymbol{\varepsilon}-\mathbf{z}_0$. The velocity predictor $\mathbf{v}_{\theta}(\mathbf{z}_t,t,\mathbf{c})$ with backbone parameters $\theta$ is trained to estimate the target velocity $\boldsymbol{\varepsilon}-\mathbf{z}_0$:
\begin{equation}
\mathcal{L}_{\mathbf{v}}(\theta)
= \mathbb{E}\!\left[\left\|(\boldsymbol{\varepsilon}-\mathbf{z}_0) - \mathbf{v}_{\theta}(\mathbf{z}_t,t,\mathbf{c})\right\|_2^2\right],
\label{FM_Loss}
\end{equation}
where the expectation is over $(\mathbf{x},\mathbf{c})\sim p_{\mathrm{data}}$, $t\sim\mathcal{U}([0,1])$, and $\boldsymbol{\varepsilon}\sim\mathcal{N}(\mathbf{0},\mathbf{I})$. Although the path is defined from data to noise, sampling proceeds in reverse from noise to data.

\begin{table}[t]
\centering
\caption{Unified formulation of denoising diffusion and flow matching under the native-prediction view.}
\label{DD_FM_Unified}
\scriptsize
\setlength{\tabcolsep}{2pt}
\renewcommand{\arraystretch}{1.3}
\begin{tabular}{@{}c@{}}
\resizebox{\columnwidth}{!}{%
\begin{tabular}{lcc}
\toprule
\textbf{Aspect} & \textbf{Diffusion} & \textbf{Flow-Matching} \\
\midrule
Noise-level convention & $\{1,\dots,T\}$ & $[0,1]$ \\
Mixing coefficients & $(\sqrt{\bar{\alpha}_t},\sqrt{1-\bar{\alpha}_t})$ & $(1-t,t)$ \\
Prediction target & $\boldsymbol{\varepsilon}$ & $\boldsymbol{\varepsilon}-\mathbf{z}_0$ \\
Backbone predictor & $\boldsymbol{\varepsilon}_{\theta}(\mathbf{z}_t,t,\mathbf{c})$ & $\mathbf{v}_{\theta}(\mathbf{z}_t,t,\mathbf{c})$ \\
\midrule
\rowcolor{gray!20} Perturbed latent & \multicolumn{2}{c}{$a_t\mathbf{z}_0+b_t\boldsymbol{\varepsilon}$} \\
\rowcolor{gray!20} Prediction error & \multicolumn{2}{c}{$\|\mathbf{u}_t-\mathbf{f}_{\theta}(\mathbf{z}_t,t,\mathbf{c})\|_2^2$} \\
\bottomrule
\end{tabular}%
}
\end{tabular}
\end{table}

\subsection{Unified Prediction Objective}
\label{Unified_Objective}

Despite different perturbation paths and prediction targets, denoising diffusion~\citep{DM} and flow matching~\citep{FM, RF} can be understood through a unified native prediction framework. As summarized in Table~\ref{DD_FM_Unified}, both paradigms construct a perturbed latent $\mathbf{z}_t = a_t\,\mathbf{z}_0 + b_t\,\boldsymbol{\varepsilon}$ with paradigm-specific mixing coefficients, and train the backbone predictor $\mathbf{f}_{\theta}(\mathbf{z}_t,t,\mathbf{c})$ to estimate the corresponding native target $\mathbf{u}_t$. Under this view, both paradigms share the unified training objective:
\begin{equation}
\theta^{\star} = \arg\min_{\theta}\;\mathbb{E}\!\left[\left\|\mathbf{u}_t - \mathbf{f}_{\theta}(\mathbf{z}_t,t,\mathbf{c})\right\|_2^2\right],
\label{DM_FM_Loss}
\end{equation}
where $\theta^{\star}$ denotes the trained backbone parameters, and the expectation is taken over the training data, noise levels, and Gaussian noise under the corresponding paradigm (Table~\ref{DD_FM_Unified}).

This unified objective motivates a source-dependent prediction consistency signal. Although Equation~\ref{DM_FM_Loss} does not guarantee that the source backbone always yields the minimum prediction error for every individual image, candidate backbones can exhibit systematically different responses to perturbed latents derived from the query image. We hypothesize that the source backbone tends to preserve stronger native prediction consistency under such perturbations, resulting in relatively lower prediction errors across informative noise levels. We validate this hypothesis in Section~\ref{Effectiveness} and exploit it as the model-level attribution signal in the fine-grained stage.

\section{Problem Formulation}
\label{Problem}
To facilitate the discussion, we first define belonging and non-belonging images, then introduce the model family structure, the attribution objective, and the corresponding access assumptions.

\subsection{Belonging vs. Non-Belonging}
Given an image generative model $M$ and image space $\mathcal{X}$, an image $\mathbf{x}\in\mathcal{X}$ is called a \emph{\textbf{belonging image}} of $M$ if it is generated by $M$, denoted by $\mathrm{src}(\mathbf{x})=M$; otherwise it is a \emph{\textbf{non-belonging image}}. Non-belonging images may originate from other known or unknown generative models, or from natural image distributions. This definition emphasizes the actual generative relationship between an image and its source model, rather than distributional overlap or visual similarity.

\subsection{Model Families and Variants}
\label{Model_Family}
Existing latent generative models~\citep{SD3, FLUX1, FLUX2} often exhibit pronounced familial relationships: a single base model or architecture gives rise to multiple variants through different training stages, fine-tuning strategies, or distillation configurations. These variants share the same VAE or highly compatible VAEs while differing in their generative backbones. Formally, given a candidate model set $\mathcal{M}=\{M_1,\dots,M_N\}$, let $V_i=(\mathcal{E}_i,\mathcal{D}_i)$ denote the VAE used by model $M_i$. We partition $\mathcal{M}$ into $K$ disjoint model families based on VAE compatibility:
\begin{equation}
\mathcal{M}=\bigcup_{k=1}^{K}\mathcal{F}_k,\quad
\mathcal{F}_k=\{M_i\mid V_i\simeq V^{(k)}\},
\label{eq:family}
\end{equation}
where $V^{(k)}$ is the representative VAE of \emph{\textbf{model family}} $\mathcal{F}_k$, and $V_i \simeq V^{(k)}$ indicates that $M_i$ uses the same or highly compatible VAE, i.e., both encode to and decode from the shared latent space with negligible distributional mismatch. Models $M_i, M_j \in \mathcal{F}_k$ are called \emph{\textbf{within-family variants}}.

\subsection{Attribution Objective}
\label{Attribution_2_Stage}

Given the candidate model set $\mathcal{M}$ and its family partition $\{\mathcal{F}_k\}_{k=1}^{K}$, the goal is to attribute a query image $\mathbf{x}$ to its source model under an open-set family-level and closed-set within-family protocol. If $\mathrm{src}(\mathbf{x})\in\mathcal{M}$, the auditor should identify its exact source model. Otherwise, when $\mathbf{x}$ originates from a source outside the candidate family set, it should be rejected as an unknown source. Unknown variants that belong to a candidate family but are not included in the enumerated within-family candidate set are outside the primary scope considered here. We decompose this objective into two stages.

\textbf{Stage~1:} Coarse-grained family-level attribution.
The first stage performs open-set family attribution, determining whether $\mathbf{x}$ can be assigned to a known model family. Let $\Phi_{\mathrm{coarse}}$ denote the coarse attribution function, with output $\hat{\mathcal{F}}=\Phi_{\mathrm{coarse}}(\mathbf{x}) \in \{\mathcal{F}_1,\dots,\mathcal{F}_K\}\cup\{\varnothing\}$, where $\hat{\mathcal{F}}$ is the predicted model family and $\varnothing$ indicates an open-set rejection label, meaning that no unique candidate-family assignment is made. If $\hat{\mathcal{F}}=\varnothing$, the attribution process terminates and no fine-grained model decision is performed.

\textbf{Stage~2:} Fine-grained model-level attribution.
When $\hat{\mathcal{F}}\neq\varnothing$, the second stage performs closed-set model attribution within the predicted family $\hat{\mathcal{F}}$. Let $\Phi_{\mathrm{fine}}$ denote the fine-grained attribution function. It outputs $\hat{M}=\Phi_{\mathrm{fine}}(\mathbf{x};\hat{\mathcal{F}})$, where $\hat{M}\in\hat{\mathcal{F}}$. Since this stage is restricted to the family selected by Stage~1, the candidate space is reduced from all models to within-family variants.

We emphasize that open-set handling is confined to Stage~1, which is designed to reject inputs outside the candidate family set. Conditional on acceptance into a candidate family, Stage~2 operates in a closed-set setting and assumes that the within-family candidate variants are known. This setting is aligned with open-weight model auditing, in which the relevant public or versioned checkpoints can be enumerated. Rejecting an unknown within-family variant constitutes a more challenging open-set problem and falls outside the primary scope of DNA.

\textbf{End-to-End.}
Let $\mathcal{F}^{\star}$ denote the true source family of $\mathbf{x}$ under the evaluation protocol: $\mathcal{F}^{\star}=\mathcal{F}_k$ if $\mathrm{src}(\mathbf{x})\in\mathcal{F}_k$, and $\mathcal{F}^{\star}=\varnothing$ if $\mathbf{x}$ originates from a source outside the candidate family set. We define the end-to-end correctness predicate $\mathrm{E2E}(\mathbf{x})$, which holds if and only if:
\begin{equation}
\mathrm{E2E}(\mathbf{x}) \Leftrightarrow
\begin{cases}
\hat{\mathcal{F}}=\mathcal{F}^{\star},\,
\hat{M}=\mathrm{src}(\mathbf{x}),&\hspace{-0.5em} \mathcal{F}^{\star}\neq\varnothing,\\
\hat{\mathcal{F}}=\varnothing, &\hspace{-0.5em} \mathcal{F}^{\star}=\varnothing.
\end{cases}
\label{eq:e2e}
\end{equation}
This criterion requires correct open-set rejection for sources outside the candidate family set and, for known-source images, correct family identification followed by exact within-family source-model attribution. Unknown variants belonging to an accepted candidate family are not explicitly rejected in the current closed-set Stage~2 setting.

\subsection{Access Assumptions}

As motivated in Section~\ref{Intro}, DNA targets open-weight auditing scenarios in which the candidate model checkpoints are publicly available to the auditor. We formalize this operating premise by specifying the access assumptions for each stage.

\textbf{Stage~1: VAE-only access.} In this stage, we assume that the auditor can access only the VAE of each candidate family, without accessing the corresponding generative backbones. This VAE-only assumption is less demanding than approaches requiring full white-box access~\citep{RONAN, LatentTracer} to the entire model. It also aligns with the family structure defined in Section~\ref{Model_Family}: since within-family variants often share the same VAE or highly compatible VAEs, maintaining one representative VAE per family is sufficient for family-level screening, reducing the maintenance cost when new families are added.

\textbf{Stage~2: Forward-only backbone access.}
In this stage, we assume that the auditor has forward-only access to the generative backbones of the variants within the predicted family $\hat{\mathcal{F}}$. The auditor can query each backbone $\mathbf{f}_{\theta}(\mathbf{z}_t,t,\mathbf{c})$ with a perturbed latent $\mathbf{z}_t$, a noise level $t$, and a condition $\mathbf{c}$, and obtain only its native prediction output; no gradient access or parameter updates are required. The query burden of this assumption is controlled by the problem structure and auditing scenarios. (i) Stage~1 has narrowed the search space from the candidate pool to a single family, so forward access is required only for the small set of within-family variants. (ii) Within-family variants often share the same development lineage and are released as checkpoints or derived from public base weights, making family-level auditing a natural unit.

Together, the two-stage access assumptions align with the attribution granularity: the coarse stage performs open-set family-level attribution with minimal VAE-level access, while the fine stage performs precise discrimination within a compact within-family candidate set using forward-only backbone queries (Section~\ref{Method_DNA}).

\section{Methodology}
\label{Method_DNA}

\begin{figure}[t]
\centering
\includegraphics[width=1\columnwidth]{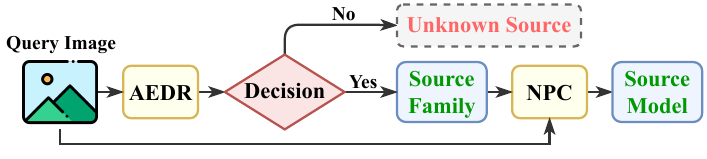}
\caption{Overview of the DNA pipeline. A query image is first processed by AEDR for open-set family-level decision, yielding either an unknown-source rejection or a predicted source family. For accepted images, NPC performs within-family attribution and outputs the final source model.}
\label{DNA_toy}
\end{figure}

\begin{figure*}[!t]
\centering
\includegraphics[width=1\textwidth]{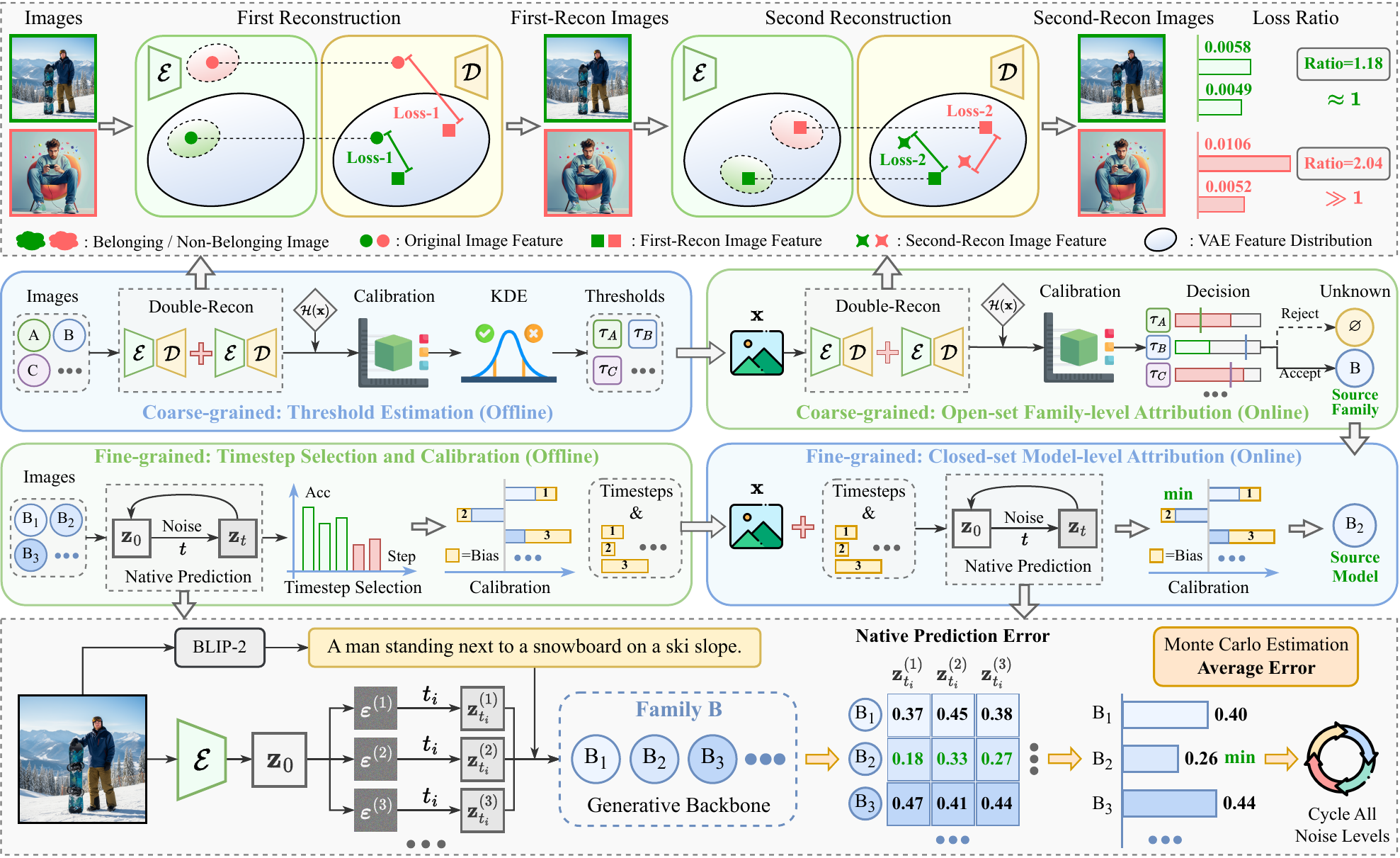}
\caption{Overview of DNA. Top: AEDR for coarse-grained family-level attribution under open-set family-level conditions, where KDE threshold estimation supports VAE double-reconstruction for family identification and rejection. Bottom: NPC for fine-grained model-level attribution within a selected family, where discriminative timestep selection and calibration enable semantically conditioned native prediction for variant discrimination.}
\label{DNA_Framework}
\end{figure*}

DNA follows the two-stage attribution framework defined in Section~\ref{Problem}, exploiting complementary signals at the VAE and backbone levels. As illustrated in Figures~\ref{DNA_toy} and~\ref{DNA_Framework}, Stage~1 extends our prior AEDR~\citep{AEDR} with a multi-family open-set decision mechanism to identify the source family, while Stage~2 introduces NPC for closed-set attribution among within-family variants under the unified native prediction objective (Section~\ref{Unified_Objective}), supporting both $\boldsymbol{\varepsilon}$-prediction and $\mathbf{v}$-prediction paradigms. DNA does not fine-tune the candidate VAEs or generative backbones and does not train an auxiliary neural network. It only uses supervised calibration: labeled samples determine family-specific thresholds, discriminative timesteps, and model-specific scalar offsets.

\subsection{Coarse-Grained Stage: AEDR}
\label{AEDR_VAE}

\subsubsection{Double-Reconstruction}
Building on the asymmetric reconstruction behavior described in Section~\ref{VAE_Preliminaries}, AEDR exploits the observation that a VAE responds differently to belonging and non-belonging images across two reconstructions. Given the VAE reconstruction operator $\mathcal{R}_k$ of candidate family $\mathcal{F}_k$ and a query image $\mathbf{x}$, we perform two cascaded reconstructions and record the corresponding losses:
\begin{equation}
\mathbf{x}_k^{(1)}=\mathcal{R}_k(\mathbf{x}),\quad
\ell_k^{(1)}=d_{\mathrm{mse}}\!\left(\mathbf{x}_k^{(1)},\mathbf{x}\right),
\label{eq:recon1}
\end{equation}
\begin{equation}
\mathbf{x}_k^{(2)}=\mathcal{R}_k\!\left(\mathbf{x}_k^{(1)}\right),\quad
\ell_k^{(2)}=d_{\mathrm{mse}}\!\left(\mathbf{x}_k^{(2)},\mathbf{x}_k^{(1)}\right),
\label{eq:recon2}
\end{equation}
where $d_{\mathrm{mse}}(\cdot,\cdot)$ denotes the mean squared error, and $\mathbf{x}_k^{(1)}$, $\mathbf{x}_k^{(2)}$ are the reconstructed images.

The key intuition is straightforward and empirically consistent. For a belonging image $\mathrm{src}(\mathbf{x})\in\mathcal{F}_k$, the image already lies near the VAE training manifold, so both reconstructions produce similar losses, i.e., $\ell_k^{(2)}(\mathbf{x})\approx\ell_k^{(1)}(\mathbf{x})$. For a non-belonging image $\mathrm{src}(\mathbf{x})\notin\mathcal{F}_k$, the first reconstruction projects it from an out-of-distribution region toward the VAE manifold with a higher loss, while the second operates on the already-projected $\mathbf{x}_k^{(1)}$ and yields a much smaller loss, i.e., $\ell_k^{(2)}(\mathbf{x})\ll\ell_k^{(1)}(\mathbf{x})$. Based on this asymmetric behavior, we define the VAE-level family attribution signal as the ratio between the two reconstruction losses:
\begin{equation}
r_k(\mathbf{x})=\frac{\ell_k^{(1)}(\mathbf{x})}{\ell_k^{(2)}(\mathbf{x})}.
\label{eq:ratio}
\end{equation}
For belonging images, $r_k(\mathbf{x})\approx 1$, while for non-belonging images $r_k(\mathbf{x})\gg 1$, providing a more stable family-level signal than a single absolute reconstruction error~\citep{RONAN, LatentTracer}.

\subsubsection{Homogeneity Calibration}
Although the double-reconstruction ratio $r_k(\mathbf{x})$ is more stable than a single absolute reconstruction error~\citep{AEDR}, its fluctuations can still be affected by the intrinsic content complexity of the image. To mitigate this bias, we introduce a homogeneity calibration based on the gray-level co-occurrence matrix (GLCM):
\begin{equation}
\mathcal{H}(\mathbf{x})=\sum_{i=0}^{Q-1}\sum_{j=0}^{Q-1}\frac{P_{\mathbf{x}}(i,j)}{1+(i-j)^2},
\label{eq:homogeneity}
\end{equation}
\begin{equation}
\tilde{r}_k(\mathbf{x}) =
\mathcal{H}(\mathbf{x})\cdot r_k(\mathbf{x}),
\label{eq:calibration}
\end{equation}
where $P_{\mathbf{x}}(i,j)$ denotes the normalized co-occurrence probability of gray levels $i$ and $j$ in image $\mathbf{x}$ under a predefined direction and offset, and $Q$ is the number of gray-level quantization bins, set to $32$ by default. A larger $\mathcal{H}(\mathbf{x})$ indicates smoother local gray-level variations and stronger content homogeneity, whereas a smaller value suggests more complex texture structures. The calibrated signal $\tilde{r}_k(\mathbf{x})$ is then used for subsequent family-level attribution decisions.

\subsubsection{Threshold Determination}
The calibrated signal $\tilde{r}_k(\mathbf{x})$ exhibits different non-parametric distributions across model families, making a single global threshold unsuitable. Therefore, we independently estimate a more robust family-specific adaptive threshold from the signal distribution of belonging images. Given $n$ belonging calibration images from family $\mathcal{F}_k$, with calibrated signals $\{\tilde{r}_{k,i}\}_{i=1}^{n}$, we estimate the empirical density using kernel density estimation (KDE)~\citep{KDE-2} and set the family threshold as the $(1-\alpha)$ quantile. Unless otherwise specified, we use $n=100$ calibration images per family.
\begin{equation}
\tau_k = \inf\!\Bigg\{\:u \,\Big|\: \int_{-\infty}^{u} \frac{1}{nh}\sum_{i=1}^{n}\mathcal{K}\:\left(\frac{y-\tilde{r}_{k,i}}{h}\right)dy \:\ge\: 1-\alpha\,\Bigg\}.
\label{eq:kde}
\end{equation}
Here, $\mathcal{K}(\cdot)$ is a Gaussian kernel, $h$ is the kernel bandwidth, and $\alpha$ is a tail tolerance parameter (default $\alpha{=}0.03$) controlling the threshold sensitivity to extreme samples. An image $\mathbf{x}$ is classified as belonging to family $\mathcal{F}_k$ if $\tilde{r}_k(\mathbf{x})\leq\tau_k$, and as non-belonging otherwise.

\subsubsection{Open-Set Family Attribution}
\label{AEDR_Open_Set}
We instantiate the coarse attribution function $\Phi_{\mathrm{coarse}}$ defined in Section~\ref{Attribution_2_Stage} using AEDR~\citep{AEDR}. A straightforward solution would be to formulate it as an open-set multi-class classifier, but this is not well aligned with the nature of AEDR signals. For each family $\mathcal{F}_k$, the calibrated signal $\tilde{r}_k(\mathbf{x})$ only measures whether $\mathbf{x}$ conforms to the VAE reconstruction behavior of that family, and the threshold $\tau_k$ defines a family-specific acceptance boundary. Therefore, $\tilde{r}_k(\mathbf{x})$ should be treated as an intrinsic family-wise decision statistic rather than a globally comparable score across families. Accordingly, we adopt $K$ parallel family-wise threshold tests. Let $\mathcal{A}(\mathbf{x})=\{\mathcal{F}_k \mid \tilde{r}_k(\mathbf{x})\leq\tau_k,\;k=1,\dots,K\}$ denote the set of families that accept $\mathbf{x}$. The coarse-grained output follows a \emph{unique-acceptance rule}:
\begin{equation}
\hat{\mathcal{F}} = \Phi_{\mathrm{coarse}}(\mathbf{x}) =
\begin{cases}
\mathcal{F}_k, & \text{if } \mathcal{A}(\mathbf{x})=\{\mathcal{F}_k\},\\[4pt]
\varnothing, & \text{if } |\mathcal{A}(\mathbf{x})|\neq 1.
\end{cases}
\label{eq:coarse-decision}
\end{equation}

Under this rule, AEDR proceeds to the fine-grained stage only when exactly one candidate family accepts $\mathbf{x}$. If no family accepts $\mathbf{x}$, or if multiple families accept it simultaneously, the coarse stage outputs the open-set rejection label $\varnothing$, indicating that no unique and reliable known family can be assigned. This parallel thresholding rule matches the family-intrinsic nature of AEDR signals and scales naturally. When a new family $\mathcal{F}_{K+1}$ is introduced, only its calibrated signal distribution and decision threshold need to be estimated, without recalibrating or modifying the existing family-wise decision rules.

\subsection{Fine-Grained Stage: NPC}
\label{NPC_Prediction}
Once a candidate family is identified, attribution is restricted to its within-family variants. Because these variants share the same VAE, VAE-level signals provide limited discrimination. We therefore propose NPC, which compares native prediction errors of candidate backbones under shared semantic conditioning and identical perturbed latent inputs, with the source backbone expected to exhibit stronger native prediction consistency (Figure~\ref{AEDR-NPC-All}), as discussed in Section~\ref{Preliminaries}.

\subsubsection{Semantic Native Prediction}
Let $\hat{\mathcal{F}}=\Phi_{\mathrm{coarse}}(\mathbf{x})$ be the predicted family from AEDR~\citep{AEDR}. When $\hat{\mathcal{F}}=\mathcal{F}_{\hat{k}}\neq\varnothing$, we re-index the within-family variants as $\mathcal{F}_{\hat{k}}=\{M_1,\dots,M_J\}$ for notational simplicity, where $J=|\mathcal{F}_{\hat{k}}|$ is the number of within-family variants. NPC first extracts a semantic condition $\mathbf{c}$ from the query image $\mathbf{x}$ using BLIP-2~\citep{BLIP2}. This condition is shared by all candidate backbones, so that the comparison of prediction errors is less affected by semantic discrepancies and more directly reflects how well each generative backbone fits the query image $\mathbf{x}$. 

The image $\mathbf{x}$ is encoded into a latent $\mathbf{z}_0=\mathcal{E}_{\hat{k}}(\mathbf{x})$ via the family-shared VAE encoder. Given a timestep set $\mathcal{T}_{\hat{k}}=\{t_1,\dots,t_L\}$, NPC samples $S$ Gaussian noise vectors $\{\boldsymbol{\varepsilon}^{(s)}\}_{s=1}^{S}\!\overset{\mathrm{i.i.d.}}\sim\!\mathcal{N}(\mathbf{0},\mathbf{I})$ at each timestep $t_i$. To ensure fair comparison, the same noise samples are shared across all candidate variants at each timestep. Following the unified formulation in Table~\ref{DD_FM_Unified}, the perturbed latent is $\mathbf{z}_{t_i}^{(s)}=a_{t_i}\mathbf{z}_0+b_{t_i}\boldsymbol{\varepsilon}^{(s)}$ and the native prediction target is $\mathbf{u}_{t_i}^{(s)}$, instantiated as $\boldsymbol{\varepsilon}^{(s)}$ for denoising diffusion and $\boldsymbol{\varepsilon}^{(s)}-\mathbf{z}_0$ for flow matching. The native prediction error of candidate variant $M_j$ at timestep $t_i$ with noise sample $s$ is:
\begin{equation}
e_{j,t_i}^{(s)} = \left\|\mathbf{u}_{t_i}^{(s)} - \mathbf{f}_{\theta_j}\!\left(\mathbf{z}_{t_i}^{(s)},\,t_i,\,\mathbf{c}\right)\right\|_2^2,
\label{eq:npc-error}
\end{equation}
where $\mathbf{f}_{\theta_j}$ is the backbone predictor of $M_j$, corresponding to $\boldsymbol{\varepsilon}_{\theta_j}$ for denoising diffusion or $\mathbf{v}_{\theta_j}$ for flow matching. Since the true expectation over $\boldsymbol{\varepsilon}$ generally has no closed-form expression, NPC approximates it via more robust Monte Carlo~\citep{Monte_Carlo} estimation with $S$ independent noise samples, reducing the variance introduced by any single perturbation:
\begin{equation}
\mu_{j,t_i} \!=\! \frac{1}{S}\sum_{s=1}^{S}e_{j,t_i}^{(s)} \!\xrightarrow{\,S\to\infty\,}\! \mathbb{E}_{\boldsymbol{\varepsilon}}\!\left[\left\|\mathbf{u}_{t_i} \!- \mathbf{f}_{\theta_j}(\mathbf{z}_{t_i},\,t_i,\,\mathbf{c})\right\|_2^2\right].
\label{eq:npc-mu}
\end{equation}
The quantity $\mu_{j,t_i}$ characterizes the native prediction consistency of candidate variant $M_j$ at noise level $t_i$ under the given semantic condition. When $\mathbf{x}$ originates from $M_j$, we empirically expect the source backbone to exhibit relatively stronger prediction consistency over informative noise levels, resulting in lower native prediction errors than competing within-family variants. This relative difference serves as the basis for the timestep selection, score normalization, and final decision.

\subsubsection{Discriminative Timesteps}
Despite their architectural similarity, within-family variants exhibit different degrees of separability in native prediction error across noise levels: some timesteps clearly distinguish the source from
non-source candidates, whereas others are considerably less discriminative (Figure~\ref{AEDR-NPC-All}). Indiscriminately aggregating all timesteps may therefore dilute the discriminative signal from the most informative ones. Accordingly, NPC uses a small labeled calibration set to select a family-specific subset of discriminative timesteps $\mathcal{T}_{\hat{k}}^{\star}\subseteq\mathcal{T}_{\hat{k}}$.

Let $\{(\mathbf{x}_i,j_i^{\star})\}_{i=1}^{n_{\mathrm{cal}}}$ be the calibration set from family $\mathcal{F}_{\hat{k}}$, comprising $n$ labeled images per variant, i.e., $n_{\mathrm{cal}}=n\times{J}$ with $n=100$ by default. Here, $j_i^{\star}\in\{1,\dots,J\}$ denotes the source-model index of $\mathbf{x}_i$. For each timestep $t\in\mathcal{T}_{\hat{k}}$, we define its single-timestep attribution accuracy $p_{\hat{k}}(t)$ as the fraction of calibration samples correctly attributed when the decision is made solely based on the native prediction errors $\mu_{j,t}(\mathbf{x}_i)$ at that timestep:
\begin{equation}
p_{\hat{k}}(t)=\frac{1}{n_{\mathrm{cal}}}
\sum_{i=1}^{n_{\mathrm{cal}}}
\mathbbm{1}\!\left[
\arg\min_{j\in[J]}\mu_{j,t}(\mathbf{x}_i)=j_i^{\star}
\right].
\label{eq:timestep-acc}
\end{equation}
The discriminative timestep set is then obtained by selecting the top-$L^{\star}$ timesteps with the highest $p_{\hat{k}}(t)$: $\mathcal{T}_{\hat{k}}^{\star}=\mathrm{Top}_{L^{\star}}(\mathcal{T}_{\hat{k}};\,p_{\hat{k}}(t))$, where $L^{\star}\leq L=|\mathcal{T}_{\hat{k}}|$. The selection is performed independently for each model family, thereby retaining the timesteps that are most informative for discriminating among its within-family variants. Empirically, timesteps closer to the noise endpoint tend to make perturbed latents approach pure Gaussian noise, substantially weakening latent-space structure and increasingly obscuring variant-specific prediction differences with random perturbations. In contrast, low-to-intermediate noise levels better preserve latent structure (see Figure~\ref{AEDR-NPC-All}), enabling NPC to reveal subtle differences among within-family variants under shared semantic conditioning. The selection therefore tends to favor these more informative and discriminative timesteps.

\subsubsection{Consistency Scores}
Raw native prediction errors are insufficient for precise attribution. First, error magnitudes vary across timesteps, allowing timesteps with larger error scales to dominate the aggregation and obscure discriminative relative differences. Second, within-family variants may exhibit variant-specific error offsets due to differences in training strategies, potentially masking the expected minimum-error advantage of the source model. To address these issues, NPC constructs consistency scores through normalization and calibration.

\textit{Cross-model z-score normalization.}
At each timestep $t_i\in\mathcal{T}_{\hat{k}}^{\star}$, we normalize the native prediction errors across the $J$ within-family variants:
\begin{equation}
\tilde{\mu}_{j,t_i} = \frac{\mu_{j,t_i}-\bar{\mu}_{\cdot,t_i}}{\sigma_{\cdot,t_i}}, \quad \bar{\mu}_{\cdot,t_i}=\frac{1}{J}\sum_{j=1}^{J}\mu_{j,t_i},
\label{eq:zscore}
\end{equation}
where $\sigma_{\cdot,t_i}$ is the standard deviation across variants at timestep $t_i$. This normalization reduces scale differences across timesteps and makes residual variant-specific biases more directly comparable. Averaging over the discriminative timestep set yields the base consistency score:
\begin{equation}
s_j(\mathbf{x}) = \frac{1}{L^{\star}}\sum_{t_i\in\mathcal{T}_{\hat{k}}^{\star}}\tilde{\mu}_{j,t_i}.
\label{eq:base-score}
\end{equation}

\textit{Model-level calibration.}
To compensate for residual systematic biases after normalization, we introduce a family-specific scalar calibration term $b_j^{(\hat{k})}$ for each candidate variant, yielding the final calibrated consistency score:
\begin{equation}
S_j(\mathbf{x}) = s_j(\mathbf{x}) + b_j^{(\hat{k})}.
\label{eq:calibrated-score}
\end{equation}
This calibration does not update any generative backbone or train an additional attribution classifier; it estimates only one scalar offset for each candidate variant within a family.

Specifically, we fix $b_1^{(\hat{k})}=0$ for a reference model and estimate the remaining $J{-}1$ scalars by maximizing the discrete, non-differentiable accuracy on the calibration set. Because the resulting parameter space is low-dimensional, we adopt a coarse-to-fine adaptive search, with an initial search interval derived from the range of base-score differences $\{s_j(\mathbf{x}_i)-s_1(\mathbf{x}_i)\}$ observed on the calibration set. The search is then iteratively refined around the current best solution. This procedure introduces negligible overhead while compensating for residual variant-specific biases.

\subsubsection{Closed-Set Variant Attribution}
Combining the preceding components, NPC performs fine-grained closed-set attribution by selecting the within-family variant with the minimum calibrated consistency score:
\begin{equation}
\hat{M} = \arg\min_{M_j\in\mathcal{F}_{\hat{k}}} S_j(\mathbf{x}) = \arg\min_{M_j\in\mathcal{F}_{\hat{k}}}\left[s_j(\mathbf{x}) + b_j^{(\hat{k})}\right].
\label{eq:npc-decision}
\end{equation}
By aggregating native prediction errors over the discriminative timesteps and applying cross-model normalization and model-level calibration, NPC produces comparable consistency scores that highlight subtle variant-specific differences, enabling reliable within-family source attribution.

\subsection{End-to-End Inference Pipeline}
\label{Overall_Framework}

Following the two-stage formulation in Section~\ref{Attribution_2_Stage}, the end-to-end execution of DNA serially combines AEDR and NPC. Given a query image $\mathbf{x}$, DNA first invokes $K$ parallel family-wise threshold tests for open-set family screening, producing a candidate family $\hat{\mathcal{F}}$. If $\hat{\mathcal{F}}=\varnothing$ (unknown or ambiguous source), the system terminates with an open-set rejection. If $\hat{\mathcal{F}}=\mathcal{F}_{\hat{k}}$, DNA proceeds to invoke NPC within that family for model-level attribution. The end-to-end decision is:
\begin{equation}
\Phi_{\mathrm{DNA}}(\mathbf{x})\hspace{-0.2em}=\hspace{-0.2em}
\begin{cases}
\Phi_{\mathrm{fine}}(\mathbf{x};\mathcal{F}_{\hat{k}}),
&\hspace{-0.3em} \Phi_{\mathrm{coarse}}(\mathbf{x})\hspace{-0.1em}=\hspace{-0.1em}\mathcal{F}_{\hat{k}},\\[3pt]
\varnothing,
&\hspace{-0.3em} \Phi_{\mathrm{coarse}}(\mathbf{x})\hspace{-0.1em}=\hspace{-0.1em}\varnothing,
\end{cases}
\label{eq:dna-decision}
\end{equation}
Here, $\Phi_{\mathrm{coarse}}$ and $\Phi_{\mathrm{fine}}$ are instantiated by AEDR and NPC, respectively. For known-source images, correctness requires both correct family identification by AEDR and exact within-family attribution by NPC. For unknown-source images outside the candidate family set, correctness requires rejection by Stage~1 without invoking NPC.

\section{Benchmark: DNA-30K}

To support systematic evaluation of fine-grained within-family attribution, we construct DNA-30K—to our knowledge, the first benchmark to jointly evaluate family-level open-set screening and closed-set attribution among within-family variants of latent generative models. It comprises 24 candidate models from six model families, spanning both denoising diffusion and flow matching. For open-set family-level evaluation, DNA-30K additionally includes 3{,}000 images generated by three non-candidate models and 3{,}000 natural images drawn equally from MS~COCO~\citep{COCO}, ImageNet~\citep{ImageNet-1K}, and LAION~\citep{LAION} as unknown sources. The unknown generators use \texttt{sd-vae-ft-ema} (PixArt-XL), a Wan-2.1-based VAE with a fine-tuned image decoder (Qwen-Image), and MoVQ (Kandinsky-3). These autoencoders are distinct from the known family VAEs and do not satisfy the compatibility criterion in Equation~\ref{eq:family}. Detailed compositions, resolutions, and configurations are reported in Table~\ref{DNA-30K}.

The data generation protocol is designed to minimize evaluation bias. All generated images are conditioned on human-annotated MS~COCO captions~\citep{COCO_Caption}, with the same caption set shared across models within each split to control semantic variation. Each source contributes 1{,}000 images, evenly divided into a 500-image calibration pool and a 500-image test set. The two splits are disjoint in image identity for all sources and in caption identity for generated sources. The calibration pool supports method-specific training, distribution fitting, support-set construction, calibration, and validation, whereas the held-out test set is reserved exclusively for final evaluation.

Except for the configuration-matched families described below, each candidate model uses its recommended default sampling configuration, reflecting how publicly released models are typically used in practice. In total, DNA-30K contains 27{,}000 generated images and 3{,}000 natural images, yielding 30{,}000 images. The benchmark will be publicly released to facilitate future research. To reduce confounding from generation settings, we match the scheduler, number of sampling steps, guidance scale, and resolution across all within-family variants in SD1.x, SD2.x, and Flux.2 (Table~\ref{DNA-30K}). This controlled setting limits the influence of the matched sampling factors on attribution performance and provides a stricter evaluation of whether NPC captures variant-specific backbone behavior rather than generation-configuration shortcuts.

\begin{table}[!t]
\centering
\caption{Composition of DNA-30K. Each family comprises multiple within-family variants; per source the 1,000 images are split evenly into a 500-image calibration set and a 500-image test set. GS indicates guidance scale.}
\label{DNA-30K}

\renewcommand{\arraystretch}{1.2}
\setlength{\tabcolsep}{2.5pt}

\begin{tabular}{@{}c@{}}
\resizebox{\columnwidth}{!}{%
\begin{tabular}{llccc}
\toprule
\textbf{Family} & \textbf{Model / Dataset} &
\textbf{Resolution} & \textbf{Steps} & \textbf{GS} \\
\midrule

\multirow{5}{*}{SD1.x}
& Stable-Diffusion-v1-1
& \multirow{5}{*}{512$\times$512}
& \multirow{5}{*}{50}
& \multirow{5}{*}{7.5} \\
& Stable-Diffusion-v1-2 & & & \\
& Stable-Diffusion-v1-3 & & & \\
& Stable-Diffusion-v1-4 & & & \\
& Stable-Diffusion-v1-5 & & & \\

\midrule
\multirow{4}{*}{SD2.x}
& Stable-Diffusion-2-base
& \multirow{4}{*}{512$\times$512}
& \multirow{4}{*}{50}
& \multirow{4}{*}{7.5} \\
& Stable-Diffusion-2-1-base & & & \\
& Stable-Diffusion-2-typography & & & \\
& Stable-Diffusion-2-cartoon-blip & & & \\

\midrule
\multirow{4}{*}{SD3.x}
& Stable-Diffusion-3-medium
& \multirow{4}{*}{1024$\times$1024}
& 28 & 7.0 \\
& Stable-Diffusion-3.5-medium & & 40 & 4.5 \\
& Stable-Diffusion-3.5-large & & 28 & 3.5 \\
& Stable-Diffusion-3.5-large-turbo & & 4 & 0 \\

\midrule
\multirow{4}{*}{SDXL}
& Stable-Diffusion-xl-base-0.9
& \multirow{4}{*}{1024$\times$1024}
& 40 & 7.5 \\
& Stable-Diffusion-xl-base-1.0 & & 40 & 7.5 \\
& SSD-1B & & 30 & 7.5 \\
& Segmind-Vega & & 25 & 9.0 \\

\midrule
\multirow{4}{*}{Flux.1}
& Flux.1-dev
& \multirow{4}{*}{1024$\times$1024}
& 50 & 3.5 \\
& Flux.1-Krea-dev & & 28 & 3.5 \\
& Flux.1-lite-8B & & 28 & 3.5 \\
& Chroma1-HD & & 40 & 3.0 \\

\midrule
\multirow{3}{*}{Flux.2}
& Flux.2-dev
& \multirow{3}{*}{1024$\times$1024}
& \multirow{3}{*}{50}
& \multirow{3}{*}{4.0} \\
& Flux.2-klein-base-4B & & & \\
& Flux.2-klein-base-9B & & & \\

\midrule
\multirow{6}{*}{Unknown}
& PixArt-XL-2-1024-MS
& 1024$\times$1024 & 20 & 4.5 \\
& Qwen-Image
& 1024$\times$1024 & 50 & 3.0 \\
& Kandinsky-3
& 1024$\times$1024 & 25 & 3.0 \\
& MS COCO (Real)
& 287--640$\times$270--640 & -- & -- \\
& LAION (Real)
& 287--640$\times$270--640 & -- & -- \\
& ImageNet (Real)
& 512$\times$512 & -- & -- \\

\bottomrule
\end{tabular}%
}
\end{tabular}
\end{table}

\section{Experiments}
\subsection{Experimental Setup}

\textbf{Models and data.}
All methods follow the DNA-30K calibration/test split. Unless otherwise specified, DNA uses 100 calibration images per family for Stage~1 threshold estimation and 100 calibration images per candidate model for Stage~2 discriminative-timestep selection and scalar score calibration. The calibration split is also used for validation and calibration-budget analyses, while the test split is reserved exclusively for evaluation.

\textbf{Baselines.}
We select five representative baselines: (1) De-Fake~\citep{DE-FAKE} trains a hybrid source classifier on CLIP image and text embeddings, using BLIP-generated captions when the original prompts are unavailable. (2) OCC-CLIP~\citep{OCC-CLIP} uses learnable prompts with a frozen CLIP encoder for few-shot one-class verification of whether an image belongs to a target generator. (3) LatentTracer~\citep{LatentTracer} initializes latent inversion with the candidate model's encoder and tests source membership through gradient-optimized reconstruction error. (4) FSD~\citep{Forensic} learns predictive filters using only real images and forms multiscale forensic self-descriptions for detection and open-set source attribution. (5) OmniDFA~\citep{OmniDFA} combines authenticity detection with few-shot source attribution, identifying generators from query embeddings and support-set prototypes. For both evaluation stages, the baselines use the following per-variant allocations from the calibration pool: OCC-CLIP uses 50 positive images and 50 real negative images; De-Fake uses all 500 images; FSD uses 400 images for source-distribution fitting and 100 for calibration; OmniDFA uses 10 support images and 490 calibration images; and LatentTracer uses all 500 images for threshold estimation. DNA uses 100 images per family for Stage~1 calibration and 100 images per candidate model for Stage~2 calibration.

\begin{table*}[t]
\centering
\caption{Family-wise open-set attribution performance across model families. Each entry reports ACC/F1/AUC (\%). For each target family, the metrics are averaged over binary tests against all remaining known families and unknown sources, including non-candidate generated and natural images. Best and second-best results are shown in \textbf{bold} and \underline{underlined}.}
\label{Stage_one}
\scriptsize
\setlength{\tabcolsep}{5pt}
\renewcommand{\arraystretch}{1.3}
\begin{tabular}{@{}c@{}}
\resizebox{\textwidth}{!}{%
\begin{tabular}{lcccccc}
\toprule
\multirow{2}{*}{\textbf{Family}} & \textbf{De-Fake} & \textbf{OCC-CLIP} & \textbf{LatentTracer} & \textbf{FSD} & \textbf{OmniDFA} & \multirow{2}{*}{\textbf{DNA}} \\
& \citep{DE-FAKE} & \citep{OCC-CLIP} & \citep{LatentTracer} & \citep{Forensic} & \citep{OmniDFA} & \\
\midrule
SD1.x & 82.06\,/\,84.81\,/\,88.12 & 60.45\,/\,71.64\,/\,63.80 & {\renewcommand{\ULthickness}{0.6pt}\uline{97.58}}\,/\,{\renewcommand{\ULthickness}{0.6pt}\uline{97.78}}\,/\,{\renewcommand{\ULthickness}{0.6pt}\uline{99.69}} & 86.99\,/\,88.75\,/\,92.53 & 74.25\,/\,71.79\,/\,91.43 & \textbf{98.79}\,/\,\textbf{98.84}\,/\,\textbf{99.74} \\
SD2.x & 72.28\,/\,70.42\,/\,81.68 & 51.47\,/\,63.66\,/\,53.40 & {\renewcommand{\ULthickness}{0.6pt}\uline{98.09}}\,/\,{\renewcommand{\ULthickness}{0.6pt}\uline{98.05}}\,/\,\textbf{99.86} & 65.18\,/\,48.29\,/\,92.13 & 72.08\,/\,63.18\,/\,90.23 & \textbf{99.04}\,/\,\textbf{98.98}\,/\,{\renewcommand{\ULthickness}{0.6pt}\uline{99.79}} \\
SD3.x & {\renewcommand{\ULthickness}{0.6pt}\uline{74.14}}\,/\,73.93\,/\,81.52 & 58.38\,/\,67.96\,/\,64.61 & 67.59\,/\,{\renewcommand{\ULthickness}{0.6pt}\uline{74.89}}\,/\,{\renewcommand{\ULthickness}{0.6pt}\uline{92.64}} & 65.88\,/\,58.37\,/\,79.78 & 72.05\,/\,63.34\,/\,87.66 & \textbf{96.20}\,/\,\textbf{95.97}\,/\,\textbf{99.36} \\
SDXL & 57.30\,/\,59.76\,/\,62.59 & 56.31\,/\,66.92\,/\,62.45 & 55.30\,/\,66.00\,/\,68.44 & 55.33\,/\,23.35\,/\,73.65 & {\renewcommand{\ULthickness}{0.6pt}\uline{79.61}}\,/\,{\renewcommand{\ULthickness}{0.6pt}\uline{76.18}}\,/\,{\renewcommand{\ULthickness}{0.6pt}\uline{91.08}} & \textbf{99.15}\,/\,\textbf{99.10}\,/\,\textbf{99.81} \\
Flux.1 & 76.02\,/\,{\renewcommand{\ULthickness}{0.6pt}\uline{77.71}}\,/\,82.93 & 59.99\,/\,69.07\,/\,65.81 & 67.79\,/\,74.71\,/\,84.36 & 62.11\,/\,47.69\,/\,74.28 & {\renewcommand{\ULthickness}{0.6pt}\uline{79.80}}\,/\,75.81\,/\,{\renewcommand{\ULthickness}{0.6pt}\uline{92.31}} & \textbf{94.37}\,/\,\textbf{94.14}\,/\,\textbf{98.13} \\
Flux.2 & 76.43\,/\,72.76\,/\,84.37 & 51.51\,/\,61.08\,/\,64.89 & 73.32\,/\,{\renewcommand{\ULthickness}{0.6pt}\uline{76.77}}\,/\,89.55 & 66.22\,/\,44.61\,/\,75.29 & {\renewcommand{\ULthickness}{0.6pt}\uline{78.54}}\,/\,68.94\,/\,{\renewcommand{\ULthickness}{0.6pt}\uline{89.83}} & \textbf{98.98}\,/\,\textbf{98.72}\,/\,\textbf{99.87} \\[2pt]
\rowcolor{gray!20}
\textbf{Avg.} & 73.04\,/\,73.23\,/\,80.20 & 56.35\,/\,66.72\,/\,62.49 & {\renewcommand{\ULthickness}{0.6pt}\uline{76.61}}\,/\,{\renewcommand{\ULthickness}{0.6pt}\uline{81.37}}\,/\,89.09 & 66.95\,/\,51.84\,/\,81.28 & 76.05\,/\,69.87\,/\,{\renewcommand{\ULthickness}{0.6pt}\uline{90.42}} & \textbf{97.76}\,/\,\textbf{97.62}\,/\,\textbf{99.45} \\
\bottomrule
\end{tabular}%
}
\end{tabular}
\end{table*}

\textbf{Evaluation metrics.}
Stage~1 comprises multiple parallel family-wise binary tests followed by a joint open-set decision. We first evaluate each family-wise test independently using Accuracy (Acc), F1 Score (F1), and AUROC (AUC), and then assess the overall family-screening accuracy under the joint open-set decision. Stage~2 is a closed-set multi-class attribution task, evaluated by accuracy and confusion matrices. Computational efficiency is measured by the average per-image inference time over multiple runs.

\textbf{Operating conditions.}
All experiments are implemented in Python~3.12.0 with PyTorch~2.9.1 and CUDA~12.8, on a server equipped with 8 NVIDIA RTX PRO 6000 GPUs. Each experiment is run independently on a single GPU.

\subsection{Attribution Performance}
\label{Effectiveness}

\begin{table}[!t]
\centering
\caption{Known-source family-level attribution accuracy (\%) under the joint open-set decision rule. Avg. denotes the unweighted mean across the six source families.}
\label{Stage_two}
\renewcommand{\arraystretch}{1.3}
\setlength{\tabcolsep}{3pt}
\begin{tabular}{@{}c@{}}
\resizebox{\columnwidth}{!}{%
\begin{tabular}{lccccccc}
\toprule
\multicolumn{1}{l}{\textbf{Method}} & \textbf{SD1.x} & \textbf{SD2.x} & \textbf{SD3.x} & \textbf{SDXL} & \textbf{Flux.1} & \textbf{Flux.2} & \textbf{Avg.} \\
\midrule
De-Fake & 10.00 & 7.80 & 13.65 & 13.65 & 16.30 & 7.33 & 11.46 \\
OCC-CLIP & 6.84 & 10.35 & 2.25 & 3.85 & 4.10 & 5.47 & 5.48 \\
LatentTracer & 8.68 & 1.80 & 40.25 & 38.05 & 22.65 & 31.67 & 23.85 \\
FSD & {\renewcommand{\ULthickness}{0.6pt}\uline{88.60}} & 33.60 & 48.05 & 13.95 & 35.25 & 33.13 & 42.10 \\
OmniDFA & 58.96 & {\renewcommand{\ULthickness}{0.6pt}\uline{49.15}} & {\renewcommand{\ULthickness}{0.6pt}\uline{49.50}} & {\renewcommand{\ULthickness}{0.6pt}\uline{67.65}} & {\renewcommand{\ULthickness}{0.6pt}\uline{64.15}} & {\renewcommand{\ULthickness}{0.6pt}\uline{59.07}} & {\renewcommand{\ULthickness}{0.6pt}\uline{58.08}} \\ [2pt]
\rowcolor{gray!20}
\textbf{DNA} & \textbf{98.76} & \textbf{97.75} & \textbf{89.85} & \textbf{89.20} & \textbf{89.15} & \textbf{88.80} & \textbf{92.25} \\
\bottomrule
\end{tabular}%
}
\end{tabular}
\end{table}

\subsubsection{Stage~1: open-set family-level}
We first evaluate each family-wise binary test independently and then assess the overall family-level attribution accuracy. Table~\ref{Stage_one} reports the per-family binary classification results, where each test distinguishes images from the target family from those originating from other known families or unknown sources. Thus, unknown-source samples are explicitly included in the Stage~1 evaluation. DNA achieves the highest ACC and F1 across all six families and the highest AUC on five families, ranking second only to LatentTracer on SD2.x in AUC, with a gap of only 0.07\%. Averaged over all six families, DNA attains 97.76\% ACC, 97.62\% F1, and 99.45\% AUC, substantially outperforming all five baselines.

Table~\ref{Stage_two} further reports the overall family-level attribution performance under the multi-family open-set decision rule (Section~\ref{AEDR_Open_Set}). DNA achieves the highest accuracy across all six families, averaging 92.25\%, and outperforms OmniDFA (58.08\%) and FSD (42.10\%) by 34.17\% and 50.15\%. Strong binary performance does not necessarily translate into reliable joint attribution: although LatentTracer slightly surpasses OmniDFA in average binary ACC (76.61\% vs.\ 76.05\%), its overall attribution accuracy drops substantially to 23.85\% compared with 58.08\% for OmniDFA. This gap reflects the stricter joint criterion, which requires correct acceptance of the true source family while simultaneously rejecting all non-source families. DNA remains robust under this criterion, demonstrating reliable family-level attribution in the open-set setting.

\begin{table}[!t]
\centering
\caption{End-to-end attribution accuracy (\%). Correct attribution requires both correct family identification and exact within-family source-model identification.}
\label{End2End}
\renewcommand{\arraystretch}{1.3}
\setlength{\tabcolsep}{3pt}
\begin{tabular}{@{}c@{}}
\resizebox{\columnwidth}{!}{%
\begin{tabular}{lccccccc}
\toprule
\multicolumn{1}{l}{\textbf{Method}} & \textbf{SD1.x} & \textbf{SD2.x} & \textbf{SD3.x} & \textbf{SDXL} & \textbf{Flux.1} & \textbf{Flux.2} & \textbf{Avg.} \\
\midrule
De-Fake & 3.36 & 7.00 & 8.70 & 7.95 & 9.00 & 4.60 & 6.77 \\
OCC-CLIP & 1.28 & 7.35 & 0.70 & 1.70 & 2.25 & 3.60 & 2.81 \\
LatentTracer & 1.84 & 0.30 & 14.20 & 8.40 & 6.20 & 10.67 & 6.93 \\
FSD & {\renewcommand{\ULthickness}{0.6pt}\uline{17.60}} & 11.20 & {\renewcommand{\ULthickness}{0.6pt}\uline{27.25}} & 5.50 & 26.65 & 15.20 & 17.23 \\
OmniDFA & 11.80 & {\renewcommand{\ULthickness}{0.6pt}\uline{19.20}} & 25.20 & {\renewcommand{\ULthickness}{0.6pt}\uline{34.25}} & {\renewcommand{\ULthickness}{0.6pt}\uline{36.00}} & {\renewcommand{\ULthickness}{0.6pt}\uline{32.27}} & {\renewcommand{\ULthickness}{0.6pt}\uline{26.45}} \\ [2pt]
\rowcolor{gray!20}
\textbf{DNA} & \textbf{93.92} & \textbf{96.40} & \textbf{86.30} & \textbf{87.05} & \textbf{88.60} & \textbf{82.40} & \textbf{89.11} \\
\bottomrule
\end{tabular}%
}
\end{tabular}
\end{table}

\begin{figure*}[!t]
\centering
\includegraphics[width=\textwidth]{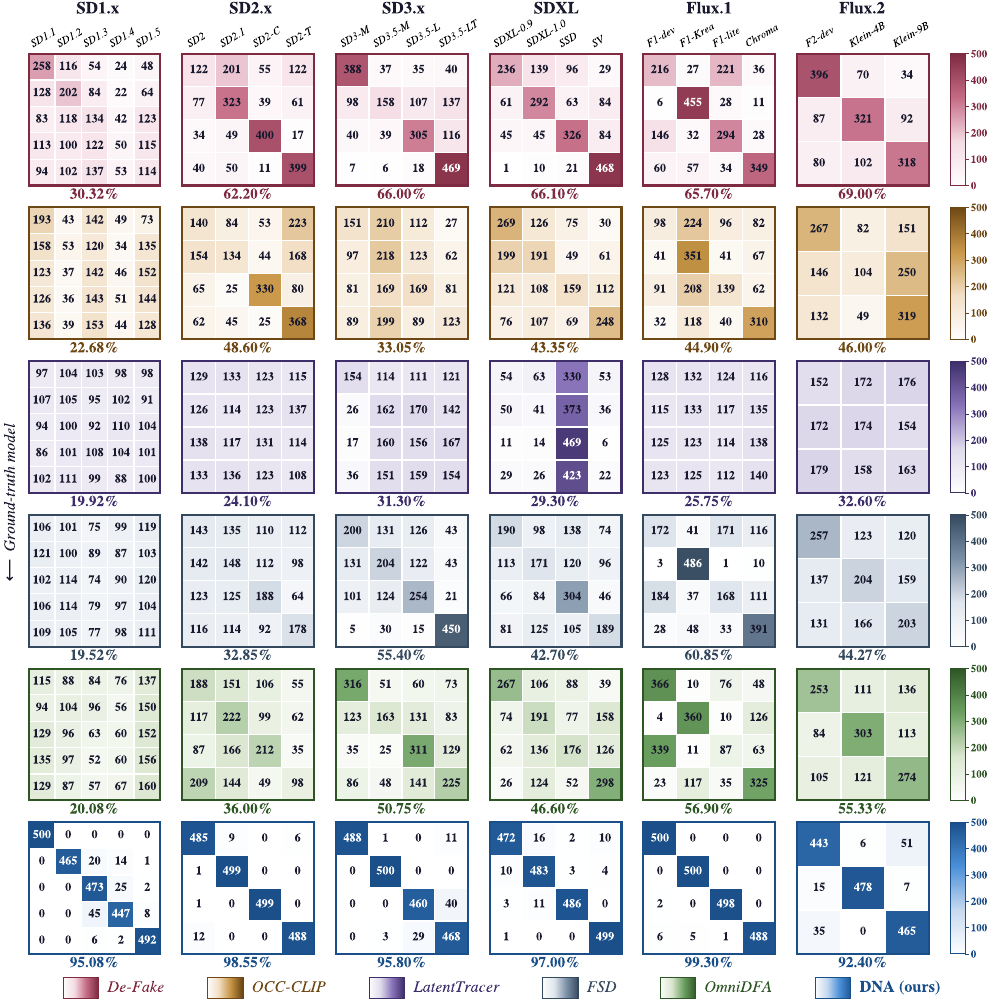}
\caption{Closed-set model-level (Stage~2) attribution confusion matrices across the six families for the six methods, under the oracle setting where Stage~1 returns the true family. Rows index the ground-truth source models, and columns index the predicted models (counts out of 500 test images per variant); per-family accuracy is annotated beneath each matrix.}
\label{Stage2_NPC}
\end{figure*}

\begin{table*}[!t]
\centering
\caption{End-to-end attribution accuracy (\%) under different combinations of Stage 1 and Stage 2 attribution methods.}
\label{Stage_Decoupling}
\scriptsize
\setlength{\tabcolsep}{5pt}
\renewcommand{\arraystretch}{1.3}
\begin{tabular}{@{}c@{}}
\resizebox{\textwidth}{!}{%
\begin{tabular}{lccccccccccc}
\toprule
\multirow{2.5}{*}{\textbf{Family}} & \multicolumn{5}{c}{\textbf{Fix Stage 1 = AEDR}} & \multicolumn{5}{c}{\textbf{Fix Stage 2 = NPC}} & \multirow{2.5}{*}{\textbf{DNA}} \\
\cmidrule(lr){2-6}\cmidrule(lr){7-11}
& De-Fake & OCC-CLIP & LatentTracer & FSD & OmniDFA & De-Fake & OCC-CLIP & LatentTracer & FSD & OmniDFA & \\
\midrule
\multicolumn{1}{l|}{SD1.x} & 29.88 & 22.52 & 19.68 & 19.32 & \multicolumn{1}{c|}{19.56} & 9.80 & 6.64 & 8.20 & {\renewcommand{\ULthickness}{0.6pt}\uline{84.40}} & \multicolumn{1}{c|}{55.88} & \textbf{93.92} \\
\multicolumn{1}{l|}{SD2.x} & {\renewcommand{\ULthickness}{0.6pt}\uline{60.95}} & 47.65 & 23.60 & 32.15 & \multicolumn{1}{c|}{35.15} & 7.80 & 10.20 & 1.75 & 33.20 & \multicolumn{1}{c|}{48.50} & \textbf{96.40} \\
\multicolumn{1}{l|}{SD3.x} & {\renewcommand{\ULthickness}{0.6pt}\uline{60.20}} & 29.95 & 28.65 & 50.90 & \multicolumn{1}{c|}{46.50} & 13.15 & 2.20 & 39.50 & 45.95 & \multicolumn{1}{c|}{47.00} & \textbf{86.30} \\
\multicolumn{1}{l|}{SDXL} & 60.50 & 38.45 & 24.30 & 37.35 & \multicolumn{1}{c|}{42.30} & 13.40 & 3.75 & 37.25 & 13.75 & \multicolumn{1}{c|}{{\renewcommand{\ULthickness}{0.6pt}\uline{65.70}}} & \textbf{87.05} \\
\multicolumn{1}{l|}{Flux.1} & 58.85 & 40.45 & 22.90 & 56.30 & \multicolumn{1}{c|}{51.15} & 16.25 & 4.10 & 22.60 & 35.20 & \multicolumn{1}{c|}{{\renewcommand{\ULthickness}{0.6pt}\uline{64.10}}} & \textbf{88.60} \\
\multicolumn{1}{l|}{Flux.2} & {\renewcommand{\ULthickness}{0.6pt}\uline{61.40}} & 41.93 & 29.00 & 39.60 & \multicolumn{1}{c|}{49.40} & 6.80 & 4.87 & 29.33 & 29.93 & \multicolumn{1}{c|}{54.60} & \textbf{82.40} \\
\midrule
\rowcolor{gray!20}
\multicolumn{1}{l|}{\textbf{Avg.}} & 55.30 & 36.83 & 24.69 & 39.27 & \multicolumn{1}{c|}{40.68} & 11.20 & 5.29 & 23.11 & 40.41 & \multicolumn{1}{c|}{{\renewcommand{\ULthickness}{0.6pt}\uline{55.96}}} & \textbf{89.11} \\
\bottomrule
\end{tabular}%
}
\end{tabular}
\end{table*}

\begin{table}[!t]
\centering
\caption{Runtime comparison (seconds) of different methods. Stage~1 runtime is shared across all families.}
\label{DNA_Run_Time}
\renewcommand{\arraystretch}{1.3}
\setlength{\tabcolsep}{3pt}
\begin{tabular}{@{}c@{}}
\resizebox{\columnwidth}{!}{%
\begin{tabular}{l|c|cccccc}
\toprule
\multirow{2}{*}{\textbf{Method}} 
& \multirow{2}{*}{\textbf{Stage 1}} 
& \multicolumn{6}{c}{\textbf{Stage 2}} \\
& & SD1.x & SD2.x & SD3.x & SDXL & Flux.1 & Flux.2 \\
\midrule
De-Fake & 1.21 & 0.16 & 0.18 & {\renewcommand{\ULthickness}{0.6pt}\uline{0.16}} & {\renewcommand{\ULthickness}{0.6pt}\uline{0.15}} & 0.16 & 0.16 \\
OCC-CLIP & 0.13 & {\renewcommand{\ULthickness}{0.6pt}\uline{0.05}} & {\renewcommand{\ULthickness}{0.6pt}\uline{0.04}} & \textbf{0.05} & \textbf{0.05} & \textbf{0.04} & \textbf{0.04} \\
LatentTracer & 117.70 & 27.28 & 21.56 & 96.90 & 139.50 & 95.62 & 71.65 \\
FSD & 2.11 & 2.14 & 2.08 & 2.12 & 2.11 & 2.10 & 2.09 \\
OmniDFA & \textbf{0.04} & \textbf{0.02} & \textbf{0.02} & \textbf{0.05} & \textbf{0.05} & {\renewcommand{\ULthickness}{0.6pt}\uline{0.05}} & {\renewcommand{\ULthickness}{0.6pt}\uline{0.05}} \\ [1pt]
\rowcolor{gray!20}
\textbf{DNA} & {\renewcommand{\ULthickness}{0.6pt}\uline{0.11}} & 15.82 & 1.03 & 32.67 & 5.27 & 28.13 & 86.68 \\
\bottomrule
\end{tabular}%
}
\end{tabular}
\end{table}

\subsubsection{Stage~2: closed-set model-level}
We evaluate NPC in isolation under an oracle setting where Stage~1 correctly identifies the source family, restricting attribution to models within the ground-truth family. Figure~\ref{Stage2_NPC} presents the confusion matrices and accuracies across all six families. DNA achieves the highest accuracy in every family, averaging 96.36\%, compared with 59.89\% for the strongest baseline, De-Fake. Specifically, DNA reaches 99.30\% on Flux.1 and maintains 95.08\% on SD1.x, the family with the most candidate variants. Several baselines remain near chance on some families and exhibit substantial confusion among variants sharing the same VAE and similar architectures, suggesting limited separability of their attribution signals. In contrast, DNA's few errors are concentrated between a small number of model pairs, consistent with the expected proximity of generative distributions among closely versioned variants. Notably, NPC achieves 95.08\%/98.55\%/92.40\% on the three configuration-matched families (SD1.x/SD2.x/Flux.2), supporting its ability to capture backbone-specific differences rather than relying solely on the matched generation settings. This finding is further supported by the sampler-robustness results in Section~\ref{Sample_Config}.

\subsubsection{End-to-End}
Table~\ref{End2End} reports end-to-end attribution accuracy, with results averaged equally across the six known-source families. DNA achieves the highest accuracy across all six families, averaging 89.11\% and surpassing the strongest baseline, OmniDFA (26.45\%), by 62.66\%. End-to-end success requires both correct family identification and correct within-family model attribution, making it more demanding than either stage alone. For some baselines, limited Stage~1 accuracy imposes a low upper bound; for instance, OCC-CLIP achieves only 5.48\% family-level accuracy. Within-family errors further constrain performance, as illustrated by OmniDFA's decrease from 58.08\% at Stage~1 to 26.45\% end-to-end. In contrast, DNA's end-to-end accuracy is only 3.14\% below its Stage~1 accuracy of 92.25\%, demonstrating strong performance across both stages. Separately, DNA achieves a 93.70\% unknown-source rejection rate, compared with 88.30\% for the best baseline (OCC-CLIP~\citep{OCC-CLIP}).

\subsubsection{Stage decoupling}
\label{Stage_decoupling}
The two stages of DNA are methodologically decoupled and can be independently substituted. To verify this modularity, we fix one stage and substitute the other with baselines. The left part of Table~\ref{Stage_Decoupling} fixes Stage~1 as AEDR, yielding gains of 14.23\%--48.53\% in average end-to-end accuracy over the original baseline pipelines. For example, De-Fake improves from 6.77\% to 55.30\%, highlighting the importance of reliable family-level attribution. The right part fixes Stage~2 as NPC, likewise improving all five baselines by 2.48\%--29.51\% in average accuracy, with OmniDFA increasing from 26.45\% to 55.96\%. With AEDR fixed as Stage~1, De-Fake is the strongest baseline Stage~2 method, reaching 55.30\%, which remains 33.81 percentage points below DNA. Across all single-stage substitutions, OmniDFA for Stage~1 combined with NPC for Stage~2 is strongest at 55.96\%; DNA exceeds this combination by 33.15 percentage points. These results demonstrate that both AEDR and NPC contribute to DNA's end-to-end advantage. The decoupled design also allows either stage to be independently upgraded without redesigning the overall framework.

\begin{table*}[!t]
\centering
\caption{Robustness to generation sampling configurations, reported as Stage~1, Stage~2, and end-to-end accuracy (\%).}
\label{Rob_Scheduler}
\small
\renewcommand{\arraystretch}{1.2}
\begin{tabular}{@{}c@{}}
\resizebox{1\textwidth}{!}{%
\begin{tabular}{lllcccc}
\toprule
\textbf{Family} & \textbf{Scheduler} & \textbf{Configuration} & \textbf{Steps} & \textbf{Stage~1} & \textbf{Stage~2} & \textbf{E2E} \\
\midrule
\multirow{5}{*}{\makecell[l]{SDXL\\{\emph{Denoising Diffusion}}}}
& \cellcolor{gray!20}EulerDiscrete (\emph{Default}) & \cellcolor{gray!20}Guidance scale (\emph{Default}) & \cellcolor{gray!20}\emph{Default} & \cellcolor{gray!20}89.20 & \cellcolor{gray!20}97.00 & \cellcolor{gray!20}\textbf{87.05} \\
& DDIM & Guidance scale = 7.5 & 50 & 90.85 & 96.15 & \textbf{87.90} \\
& EulerDiscrete & Guidance scale = 7.5 & 30 & 89.20 & 96.75 & \textbf{86.85} \\
& EulerAncestralDiscrete & Guidance scale = 7.5 & 30 & 89.75 & 94.55 & \textbf{86.35} \\
& DPMSolverMultistep & Guidance scale = 7.5, Karras $\sigma$ & 25 & 86.90 & 96.50 & \textbf{84.25} \\
\midrule
\multirow{4}{*}{\makecell[l]{Flux.1\\{\emph{Flow Matching}}}}
& \cellcolor{gray!20}FlowMatchEuler (\emph{Default}) & \cellcolor{gray!20}Guidance scale, Dynamic shift (\emph{Default}) & \cellcolor{gray!20}\emph{Default} & \cellcolor{gray!20}89.15 & \cellcolor{gray!20}99.30 & \cellcolor{gray!20}\textbf{88.60} \\
& FlowMatchEuler & Guidance scale = 3.5, Fixed shift = 3.0 & 30 & 88.20 & 99.10 & \textbf{87.55} \\
& FlowMatchEuler + Beta $\sigma$ & Guidance scale = 3.5, Fixed shift = 3.0 & 30 & 91.70 & 99.50 & \textbf{91.30} \\
& FlowMatchEuler + Karras $\sigma$ & Guidance scale = 3.5, Fixed shift = 3.0 & 30 & 88.60 & 98.40 & \textbf{87.45} \\
\bottomrule
\end{tabular}%
}
\end{tabular}
\end{table*}

\subsection{Efficiency}
\label{Efficiency}
Table~\ref{DNA_Run_Time} reports per-image inference time on DNA-30K. Stage~1 takes 0.11\,s in total to screen all six candidate families, second only to OmniDFA (0.04\,s), using two reconstructions per candidate VAE without backbone access or gradient optimization. Stage~2 takes 1.03--86.68\,s, exceeding the cost of fast feature-based baselines but achieving 96.36\% average accuracy, compared with 59.89\% for the strongest baseline, De-Fake. DNA is also faster than LatentTracer~\citep{LatentTracer} on five families in Stage~2 and all six end-to-end, while substantially improving Stage~2 accuracy (96.36\% vs.\ 27.16\%). Unlike baselines using learned prompts, classifiers, or representations~\citep{DE-FAKE,OCC-CLIP,OmniDFA}, DNA requires calibration but no auxiliary network training. NPC's computational budget can be adjusted by reducing the timestep count $L^{\star}$ or noise-sample count $S$, enabling an explicit accuracy-efficiency trade-off (Figure~\ref{Ab_Noise_Step}).

\subsection{Robustness}

\subsubsection{Sampling configurations}
\label{Sample_Config}
To verify that DNA is robust to changes in generation-time sampling configurations, we conduct experiments on two representative families: SDXL (Denoising diffusion) and Flux.1 (Flow Matching). For SDXL, we replace the default EulerDiscrete scheduler with DDIM, EulerAncestralDiscrete, and DPMSolverMultistep. For Flux.1, since the Rectified Flow path is tightly coupled with the FlowMatchEuler solver, we instead vary key configurations within this framework, including dynamic/fixed shift and $\sigma$-schedule forms (linear, Beta, Karras). As shown in Table~\ref{Rob_Scheduler}, SDXL achieves end-to-end accuracy of 84.25--87.90\% across four schedulers, while Flux.1 achieves 87.45--91.30\% across three alternative configurations, with deviations from the default within 3\%. Moreover, Stage~2 accuracy remains at least 94.55\% in all cases, suggesting that NPC primarily captures prediction consistency induced by the training objective and backbone parameters rather than specific sampling configurations.

\begin{table}[!t]
\centering
\caption{End-to-end accuracy (\%) under different calibration budgets $B$. Stage~1 uses $B$ images per family, while Stage~2 uses $B$ images per candidate model.}
\label{Rob_Samples}
\scriptsize
\renewcommand{\arraystretch}{1.2}
\setlength{\tabcolsep}{4pt}
\begin{tabular}{@{}c@{}}
\resizebox{\columnwidth}{!}{%
\begin{tabular}{lccccccc}
\toprule
\multirow{2.5}{*}{\textbf{Family}} & \multicolumn{7}{c}{\textbf{Calibration budget $B$}} \\
\cmidrule(l{1pt}r{1pt}){2-8}
& 100 & 80 & 60 & 40 & 20 & 10 & 5 \\
\midrule
SD1.x  & 93.92 & 92.20 & 92.76 & 93.12 & 89.44 & 88.56 & 87.92 \\
SD2.x  & 96.40 & 95.15 & 96.90 & 95.60 & 94.65 & 89.25 & 90.25 \\
SD3.x  & 86.30 & 86.15 & 86.40 & 87.90 & 85.10 & 77.40 & 82.80 \\
SDXL   & 87.05 & 77.00 & 77.75 & 77.90 & 73.70 & 72.25 & 73.35 \\
Flux.1 & 88.60 & 91.10 & 90.75 & 88.85 & 86.00 & 84.25 & 81.30 \\
Flux.2 & 82.40 & 80.27 & 81.87 & 82.33 & 74.47 & 70.53 & 83.87 \\
\midrule
\rowcolor{gray!20}
\textbf{Avg.} & \textbf{89.11}
& \textbf{\makecell{86.98}}
& \textbf{\makecell{87.74}}
& \textbf{\makecell{87.62}}
& \textbf{\makecell{83.89}}
& \textbf{\makecell{80.37}}
& \textbf{\makecell{83.25}} \\
\bottomrule
\end{tabular}%
}
\end{tabular}
\end{table}

\begin{figure*}[t]
\centering
\includegraphics[width=0.99\textwidth]{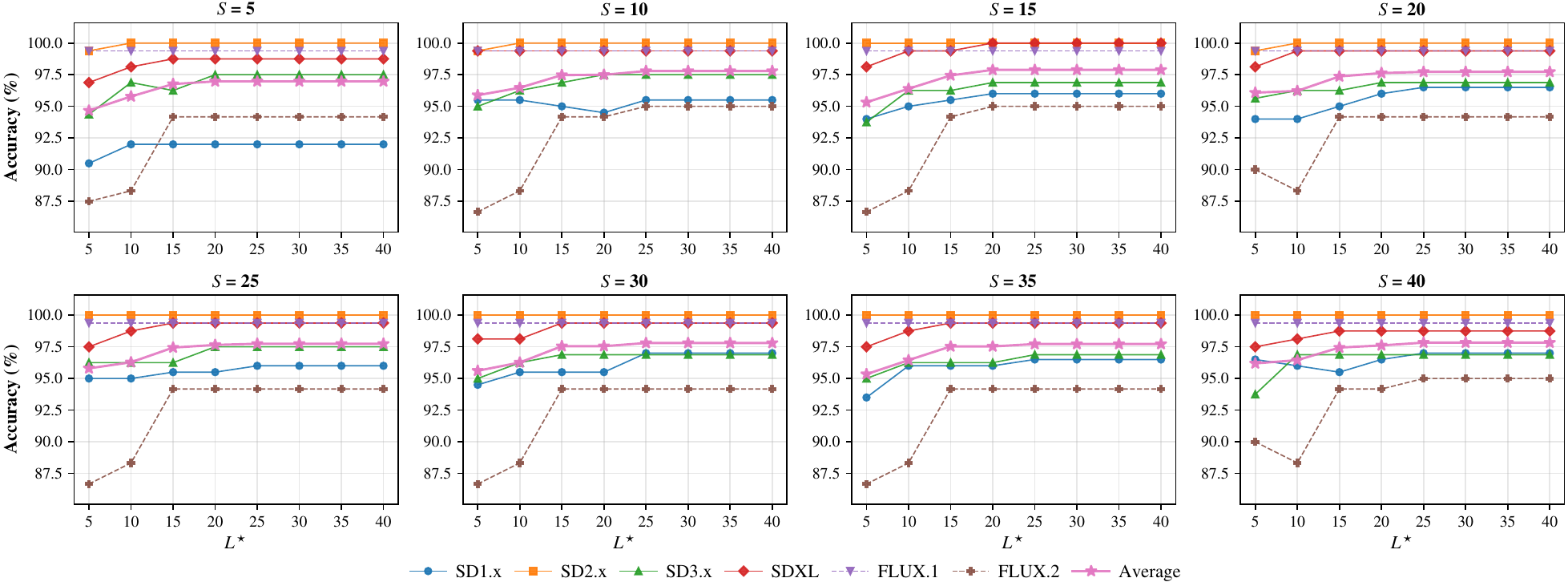}
\caption{Stage~2 accuracy over the grid of noise-sample count $S$ (panels) and discriminative timestep count $L^{\star}$ (x-axis) on the validation split. Accuracy generally saturates beyond $S,L^{\star}\geq 15$, indicating low sensitivity to these hyperparameters.}
\label{Ab_Noise_Step}
\end{figure*}

\begin{table*}[!t]
\centering
\caption{Robustness to real-world post-processing. Results are reported as Stage~1 (S1), Stage~2 (S2), and end-to-end (E2E) attribution accuracy (\%). Each platform trial uses 100 randomly selected SD2.x images.}
\label{Rob_Post_Processing}
\scriptsize
\setlength{\tabcolsep}{6pt}
\renewcommand{\arraystretch}{1.2}
\begin{tabular}{@{}c@{}}
\resizebox{\textwidth}{!}{%
\begin{tabular}{lccccccccccccccc}
\toprule
\multirow{2.5}{*}{\textbf{Method}}
& \multicolumn{3}{c}{\textbf{Clean}}
& \multicolumn{3}{c}{\textbf{JPEG}}
& \multicolumn{3}{c}{\textbf{Facebook}}
& \multicolumn{3}{c}{\textbf{Telegram}}
& \multicolumn{3}{c}{\textbf{WeChat}} \\
\cmidrule(lr){2-4}\cmidrule(lr){5-7}\cmidrule(lr){8-10}\cmidrule(lr){11-13}\cmidrule(lr){14-16}
& S1 & S2 & E2E
& S1 & S2 & E2E
& S1 & S2 & E2E
& S1 & S2 & E2E
& S1 & S2 & E2E \\
\midrule
\multicolumn{1}{l|}{De-Fake~\citep{DE-FAKE}} & 7.80 & \uline{62.20} & \multicolumn{1}{c|}{7.00} & 8.95 & \uline{56.65} & \multicolumn{1}{c|}{7.70} & 2 & \uline{53} & \multicolumn{1}{c|}{1} & 5 & 41 & \multicolumn{1}{c|}{2} & 2 & 47 & 0 \\
\multicolumn{1}{l|}{OCC-CLIP~\citep{OCC-CLIP}} & 10.35 & 48.60 & \multicolumn{1}{c|}{7.35} & 6.30 & 45.90 & \multicolumn{1}{c|}{4.60} & 6 & 25 & \multicolumn{1}{c|}{1} & 5 & 24 & \multicolumn{1}{c|}{2} & 1 & 21 & 0 \\
\multicolumn{1}{l|}{LatentTracer~\citep{LatentTracer}} & 1.80 & 24.10 & \multicolumn{1}{c|}{0.30} & 0.00 & 24.75 & \multicolumn{1}{c|}{0.00} & 7 & 20 & \multicolumn{1}{c|}{1} & 0 & 36 & \multicolumn{1}{c|}{0} & 4 & 28 & 2 \\
\multicolumn{1}{l|}{FSD~\citep{Forensic}} & 33.60 & 32.85 & \multicolumn{1}{c|}{11.20} & \uline{44.90} & 30.55 & \multicolumn{1}{c|}{\uline{15.25}} & 38 & 28 & \multicolumn{1}{c|}{5} & 9 & 34 & \multicolumn{1}{c|}{4} & \uline{39} & 24 & \uline{6} \\
\multicolumn{1}{l|}{OmniDFA~\citep{OmniDFA}} & \uline{49.15} & 36.00 & \multicolumn{1}{c|}{\uline{19.20}} & 36.10 & 32.10 & \multicolumn{1}{c|}{12.65} & 43 & 41 & \multicolumn{1}{c|}{6} & \uline{21} & \uline{62} & \multicolumn{1}{c|}{5} & 13 & \uline{64} & 2 \\
\midrule
\rowcolor{gray!20}
\multicolumn{1}{l|}{\textbf{DNA}} & \textbf{97.75} & \textbf{98.55} & \multicolumn{1}{c|}{\textbf{96.40}} & 9.55 & \textbf{82.05} & \multicolumn{1}{c|}{9.15} & \uline{55} & \textbf{99} & \multicolumn{1}{c|}{\uline{55}} & 20 & \textbf{97} & \multicolumn{1}{c|}{\uline{20}} & 8 & \textbf{81} & \uline{6} \\
\rowcolor{gray!20}
\multicolumn{1}{l|}{\textbf{DNA-R}} & \textbf{97.75} & \textbf{98.55} & \multicolumn{1}{c|}{\textbf{96.40}} & \textbf{90.60} & \textbf{82.05} & \multicolumn{1}{c|}{\textbf{75.35}} & \textbf{85} & \textbf{99} & \multicolumn{1}{c|}{\textbf{84}} & \textbf{95} & \textbf{97} & \multicolumn{1}{c|}{\textbf{92}} & \textbf{68} & \textbf{81} & \textbf{55} \\
\bottomrule
\end{tabular}%
}
\end{tabular}
\end{table*}

\subsubsection{Limited calibration samples}
We reduce the calibration budget from the default 100 samples to 5. We estimate the Stage~1 thresholds, Stage~2 discriminative timesteps, and normalized calibrated scores, and evaluate end-to-end performance on the test set, as shown in Table~\ref{Rob_Samples}. When the sample count decreases from 100 to 40, the average accuracy drops by only 1.49\% (89.11\%$\to$87.62\%). Even under the extreme 5-sample setting, end-to-end accuracy remains at 83.25\%. Performance is not strictly monotonic with the sample count, which may arise from variations in calibration-sample composition and their effects on threshold estimation and timestep selection. Nevertheless, the overall trend is stable, indicating that DNA is not overly sensitive to calibration-set size and maintains useful performance under limited annotation budgets.

\subsubsection{Real-world settings}
JPEG compression is widely used in image dissemination, motivating our robustness evaluation on SD2.x using simulated JPEG compression (QF=75) and exploratory real-world trials involving upload--download processing on Facebook, Telegram, and WeChat. Table~\ref{Rob_Post_Processing} identifies Stage~1 as DNA's primary post-processing bottleneck. We introduce DNA-R as an optional deployment extension and evaluate it on SD2.x. Self-contained lossy JPEG files include the quantization tables required for decoding, providing an explicit bitstream cue to activate the robust branch~\citep{JPEG_T81}. This branch applies pretrained PromptCIR~\citep{PromptCIR} restoration followed by the unchanged six-family AEDR tests. An SD2-specific Gaussian gate is invoked only when all six families are rejected. It uses two reconstruction errors and texture homogeneity from the SD2 VAE response to the original JPEG. After log transformation and standardization, these features are scored by the negative log-likelihood of a single Gaussian fitted on 100 SD2 positive samples with simulated QF75 compression. For both DNA and DNA-R, Stage~2 operates on the compressed input image. Under simulated JPEG, DNA-R improves Stage~1 accuracy from 9.55\% to 90.60\% and end-to-end accuracy from 9.15\% to 75.35\%, while leaving Stage~2 unchanged at 82.05\% and preserving clean performance. Without platform-specific recalibration, DNA-R achieves end-to-end accuracies of 84\%, 92\%, and 55\% on Facebook, Telegram, and WeChat, respectively, compared with 55\%, 20\%, and 6\% for DNA. These gains incur only a slight increase in false acceptance: non-SD2 known-source misattributions rise from 4/500 to 10/500, while false acceptance of real images increases only from 0/100 to 1/100.

\begin{table}[!t]
\centering
\caption{Selected NPC hyperparameters ($L^{\star}$, $S$) and Stage~2 accuracy on calibration, validation, and test sets. Parentheses show test-minus-validation differences.}
\label{Final_Noise_Step}
\scriptsize
\renewcommand{\arraystretch}{1.3}
\setlength{\tabcolsep}{4pt}
\begin{tabular}{@{}c@{}}
\resizebox{\columnwidth}{!}{%
\begin{tabular}{lccccc}
\toprule
\textbf{Family} & \textbf{$L^{\star}$} & \textbf{$S$} & \textbf{Cal Acc (\%)} & \textbf{Val Acc (\%)} & \textbf{Test Acc (\%)} \\
\midrule
SD1.x  & 25 & 30 & 96.00  & 97.00  & \makecell{95.08 (-1.92)} \\
SD2.x  & 10  & 5 & 99.17  & 100.00 & \makecell{98.55 (-1.45)} \\
SD3.x  & 20  & 5 & 95.83  & 97.50  & \makecell{95.80 (-1.70)} \\
SDXL   & 5 & 10  & 98.75  & 99.38  & \makecell{97.00 (-2.38)} \\
FLUX.1 & 5  & 5  & 100.00 & 99.38  & \makecell{99.30 (-0.08)} \\
FLUX.2 & 25  & 5 & 93.89  & 94.17  & \makecell{92.40 (-1.77)} \\
\midrule
\rowcolor{gray!20}
\textbf{Avg.} & \textbf{--} & \textbf{--} & \textbf{97.27} & \textbf{97.91} & \textbf{\makecell{96.36 (-1.55)}} \\
\bottomrule
\end{tabular}%
}
\end{tabular}
\end{table}

\subsection{Ablation Studies}
\label{sec:ablation}

\subsubsection{Noise samples and timesteps}
NPC involves two key hyperparameters: the number of noise samples $S$ and the number of discriminative timesteps $L^{\star}$. We split the 100 calibration images per model into a 60/40 calibration/validation partition and evaluate Stage~2 accuracy on the validation set over a two-dimensional grid $S, L^{\star}\in\{5,10,\dots,40\}$, as shown in Figure~\ref{Ab_Noise_Step}. Overall, the average accuracy ranges from 94.67\% to 97.88\%, indicating that NPC is not sensitive to either hyperparameter. Per-family results show that SD2.x and Flux.1 are the most stable, with accuracy consistently above 99\%. Flux.2 is relatively more sensitive, dropping to approximately 86.67\% at low $S$ and low $L^{\star}$, but recovering to about 95\% as both increase. The general trend indicates that accuracy saturates beyond $S,L^{\star}\geq 15$, with further increases primarily adding linear inference overhead. Based on the performance--efficiency trade-off, the final per-family configurations are listed in Table~\ref{Final_Noise_Step}. After transferring the parameters selected on the validation set to the test set, the average accuracy decreases from 97.91\% to 96.36\%, with the largest per-family drop not exceeding 2.38\%.

\subsubsection{Component ablation}
To assess the contribution of each component in DNA, we incrementally add five core components following their actual order of use in the framework and report the end-to-end attribution accuracy gain at each step, as shown in Table~\ref{Ab_Component}. The Base configuration uses the uncalibrated AEDR ratio for Stage~1 and raw native prediction-error aggregation for Stage~2, achieving an average end-to-end accuracy of 45.22\%. On top of this, homogeneity calibration (+Hom.), shared noise perturbation (+Shared), semantic conditioning (+Sem.), discriminative timestep selection (+DisStep), and normalized calibrated scoring (+Norm.) yield gains of 5.43\%, 6.31\%, 20.31\%, 5.09\%, and 6.76\%, respectively, ultimately raising the accuracy to 89.11\%—a cumulative improvement of 43.89\%. Semantic conditioning yields the largest incremental gain (+20.31\%), indicating that aligning the conditioning input with the generation-stage semantics allows prediction error differences to more accurately reflect distributional shifts among candidate models. The full component configuration yields consistent improvements across all families, suggesting that the proposed components are not tied to the statistical characteristics of any single model family.

\begin{table}[t]
\centering
\caption{Component-wise ablation. Intermediate columns report incremental accuracy gains, and the final column reports end-to-end attribution accuracy (\%).}
\label{Ab_Component}
\scriptsize
\renewcommand{\arraystretch}{1.3}
\setlength{\tabcolsep}{3pt}
\begin{tabular}{@{}c@{}}
\resizebox{\columnwidth}{!}{%
\begin{tabular}{lccccccc}
\toprule
\textbf{Family} & Base & +Hom. & +Shared & +Sem. & +DisStep & +Norm. & \textbf{Final} \\
\midrule
SD1.x  & 50.40 & +0.12  & +12.40 & +26.72 & +2.52  & +1.76 & \textbf{93.92} \\
SD2.x  & 54.95 & +2.90  & +5.95  & +25.55 & +0.90  & +6.15 & \textbf{96.40} \\
SD3.x  & 50.95 & +2.15  & +4.80  & +11.85 & +8.85  & +7.70 & \textbf{86.30} \\
SDXL   & 34.30 & +5.35  & +8.05  & +26.10 & +4.40  & +8.85 & \textbf{87.05} \\
Flux.1 & 43.20 & +13.30 & +1.65  & +19.65 & +3.25  & +7.55 & \textbf{88.60} \\
Flux.2 & 37.53 & +8.74  & +5.00  & +12.00 & +10.60 & +8.53 & \textbf{82.40} \\
\midrule
\rowcolor{gray!20}
\textbf{Avg.} & \textbf{45.22} & \textbf{+5.43} & \textbf{+6.31} & \textbf{+20.31} & \textbf{+5.09} & \textbf{+6.76} & \textbf{89.11} \\
\bottomrule
\end{tabular}%
}
\end{tabular}
\end{table}

\begin{table}[t]
\centering
\caption{Stage~2 accuracy (\%) under four conditioning strategies and a random baseline. \emph{Original}: ground-truth prompt; \emph{Semantic}: BLIP-2 caption; \emph{Generic}: ``an image''; \emph{None}: unconditional.}
\label{Ab_Semantic}
\scriptsize
\renewcommand{\arraystretch}{1.3}
\setlength{\tabcolsep}{3pt}
\begin{tabular}{@{}c@{}}
\resizebox{\columnwidth}{!}{%
\begin{tabular}{lcccccc}
\toprule
\textbf{Family} & \textbf{$J$} & \textbf{Original} & \textbf{Semantic} & \textbf{Generic} & \textbf{None} & \textbf{Random} \\
\midrule
SD1.x  & 5 & 97.16 & 95.08 & 73.60 & 65.52 & 20.00 \\
FLUX.1 & 4 & 99.75 & 99.30 & 91.80 & 88.40 & 25.00 \\
\bottomrule
\end{tabular}%
}
\end{tabular}
\end{table}

To examine what drives the large gain from semantic conditioning (+20.31\% in Table~\ref{Ab_Component}), we evaluate Stage~2 accuracy under four conditioning strategies on SD1.x (denoising diffusion) and FLUX.1 (flow matching), as shown in Table~\ref{Ab_Semantic}. Both families exhibit a consistent ordering: Original $\approx$ Semantic $\gg$ Generic $>$ None $\gg$ random. Original and Semantic differ by less than 2.1\% on both families, suggesting that BLIP-2 captions provide an effective deployment-time proxy for the true generation prompt. Crucially, None still substantially exceeds random guessing (65.52\% vs.\ 20\% for SD1.x; 88.40\% vs.\ 25\% for FLUX.1), suggesting that useful attribution information remains in backbone responses without image-specific conditioning. Semantic conditioning further improves Stage~2 accuracy by 29.56 percentage points on SD1.x and 10.90 percentage points on FLUX.1, supporting its role in strengthening this existing signal.

\subsubsection{Family-count scaling}
Due to the two-stage decoupling (Section~\ref{Stage_decoupling}), increasing the number of candidate families $K$ affects only Stage~1. We vary $K$ from 1 to 6 and enumerate all $\binom{6}{K}$ family subsets for each $K$, recording the Stage~1 accuracy under each subset, as shown in Table~\ref{Add_Family}. As $K$ increases, overall attribution accuracy decreases moderately. SD1.x and SD2.x are the most stable, consistently maintaining accuracy above 97\% with drops of at most 1.1\%. Other families are more sensitive to increasing $K$, with drops of approximately 5\%-10\%. Even in the most challenging $K{=}6$ setting, all six families maintain accuracy above 88\%.

\begin{table}[t]
\centering
\caption{Attribution accuracy (\%) under different values of $K$. $\dagger$ indicates that only one selection is possible. Results are reported as mean $\pm$ standard deviation.}
\label{Add_Family}
\scriptsize
\renewcommand{\arraystretch}{1.3}
\setlength{\tabcolsep}{3pt}
\begin{tabular}{@{}c@{}}
\resizebox{\columnwidth}{!}{%
\begin{tabular}{lcccccc}
\toprule
\textbf{Family} & \textbf{$K=1$} & \textbf{$K=2$} & \textbf{$K=3$} & \textbf{$K=4$} & \textbf{$K=5$} & \textbf{$K=6$} \\
\midrule
SD1.x  & 99.16$^{\dagger}$ & 99.08$\pm$0.04 & 99.00$\pm$0.05 & 98.92$\pm$0.05 & 98.84$\pm$0.04 & 98.76$^{\dagger}$ \\
SD2.x  & 98.85$^{\dagger}$ & 98.59$\pm$0.31 & 98.36$\pm$0.35 & 98.15$\pm$0.34 & 97.95$\pm$0.28 & 97.75$^{\dagger}$ \\
SD3.x  & 95.75$^{\dagger}$ & 94.30$\pm$1.05 & 93.01$\pm$1.22 & 91.85$\pm$1.16 & 90.80$\pm$0.92 & 89.85$^{\dagger}$ \\
SDXL   & 99.45$^{\dagger}$ & 96.88$\pm$3.35 & 94.63$\pm$3.83 & 92.63$\pm$3.60 & 90.83$\pm$2.79 & 89.20$^{\dagger}$ \\
Flux.1 & 93.45$^{\dagger}$ & 92.43$\pm$1.08 & 91.53$\pm$1.28 & 90.69$\pm$1.25 & 89.90$\pm$1.00 & 89.15$^{\dagger}$ \\
Flux.2 & 98.47$^{\dagger}$ & 95.76$\pm$2.08 & 93.57$\pm$2.24 & 91.74$\pm$2.01 & 90.16$\pm$1.46 & 88.80$^{\dagger}$ \\
\bottomrule
\end{tabular}%
}
\end{tabular}
\end{table}

\subsubsection{Within-family variant scaling}
Increasing the number of within-family variants $J$ affects only Stage~2. We progressively increase $J$ from 1 to the maximum family size, enumerate all possible model subsets of size $J$, and report the mean Stage~2 accuracy, as shown in Table~\ref{Add_variants}. For $J=1$, attribution is trivial and the cross-model normalization in Equation~\ref{eq:zscore} is bypassed; the sole candidate is returned directly, yielding 100\% accuracy by definition. Accuracy remains consistently high as $J$ grows, although attribution naturally becomes more challenging with more candidate variants: SD2.x and Flux.1 retain 98.55\% and 99.30\% at their maximum sizes, respectively. Flux.2 shows the largest degradation as $J$ increases, yet still achieves 92.40\% with all three candidate variants. Both scaling experiments confirm that DNA maintains stable attribution performance as the overall candidate space expands along both the inter-family and intra-family dimensions.

\begin{table}[t]
\centering
\caption{Attribution accuracy (\%) under different values of $J$. $\dagger$ indicates that only one selection is possible. Results are reported as mean $\pm$ standard deviation.}
\label{Add_variants}
\scriptsize
\renewcommand{\arraystretch}{1.3}
\setlength{\tabcolsep}{3pt}
\begin{tabular}{@{}c@{}}
\resizebox{\columnwidth}{!}{%
\begin{tabular}{lcccccc}
\toprule
\textbf{Family} & \textbf{$|\mathcal{F}_k|$} & \textbf{$J=1$} & \textbf{$J=2$} & \textbf{$J=3$} & \textbf{$J=4$} & \textbf{$J=5$} \\
\midrule
SD1.x  & 5 & 100.00$\pm$0.00 & 98.54$\pm$2.13 & 97.25$\pm$2.24 & 96.11$\pm$1.80 & 95.08$^{\dagger}$ \\
SD2.x  & 4 & 100.00$\pm$0.00 & 99.47$\pm$0.70 & 98.98$\pm$0.68 & 98.55$^{\dagger}$ & -- \\
SD3.x  & 4 & 100.00$\pm$0.00 & 98.53$\pm$2.46 & 97.15$\pm$2.24 & 95.80$^{\dagger}$ & -- \\
SDXL   & 4 & 100.00$\pm$0.00 & 98.62$\pm$0.82 & 97.75$\pm$0.77 & 97.00$^{\dagger}$ & -- \\
Flux.1 & 4 & 100.00$\pm$0.00 & 99.53$\pm$0.41 & 99.37$\pm$0.29 & 99.30$^{\dagger}$ & -- \\
Flux.2 & 3 & 100.00$\pm$0.00 & 94.87$\pm$3.39 & 92.40$^{\dagger}$ & -- & -- \\
\bottomrule
\end{tabular}%
}
\end{tabular}
\end{table}

\section{Limitations and Future Work}
DNA currently performs open-set rejection at the family level, while Stage~2 assumes that the candidate variants within an accepted family are enumerated. Extending NPC to reject unseen within-family variants therefore represents a natural next step toward fully open-set model attribution. In addition, although DNA-R substantially improves robustness to JPEG compression on SD2.x, extending such post-processing robustness consistently across diverse model families and real-world dissemination platforms remains an important direction for future work.

\section{Conclusion}
This paper proposes Dual-stage Native Attribution (DNA), a coarse-to-fine framework that decouples source-model attribution along the ``\emph{VAE}$\rightarrow$\emph{backbone}'' hierarchy into open-set family-level screening and closed-set model-level attribution. The fine-grained stage introduces Native Prediction Consistency (NPC), which supports both denoising diffusion and flow matching. We construct DNA-30K, comprising 30,000 images from 24 candidate models across six families, together with unknown-source images for open-set family-level evaluation. On DNA-30K, DNA achieves 92.25\% Stage~1 accuracy, 96.36\% Stage~2 accuracy under the oracle true-family setting, and 89.11\% end-to-end attribution accuracy, outperforming the strongest baseline by 33.15\%. DNA remains stable under candidate-set expansion, diverse sampling configurations, and limited calibration data, retaining 83.25\% end-to-end accuracy with only five calibration images per family for Stage~1 and per candidate model for Stage~2.

\section*{Statements and Declarations}

\subsection*{Funding}
This work was supported in part by the New Generation Artificial Intelligence-National Science and Technology Major Project (No. 2025ZD0123202) and by the National Natural Science Foundation of China (Grants U2336206, 62472398 and U2436601).

\subsection*{Competing interests}
The authors have no relevant financial or non-financial interests to disclose.

\subsection*{Data availability}
All synthetic images in DNA-30K were generated by the authors and will be made publicly available upon acceptance of this manuscript. The natural images were obtained from the publicly available MS COCO~\citep{COCO}, ImageNet~\citep{ImageNet}, and LAION~\citep{LAION}, subject to their respective access conditions.

\subsection*{Code availability}
The core implementation of DNA is publicly available at \url{https://github.com/wangchao0708/DNA}.








\bibliography{Ref}

\end{document}